\documentclass[11pt]{article}

\usepackage[final]{acl}

\usepackage{times}
\usepackage{latexsym}

\usepackage[T1]{fontenc}

\usepackage[utf8]{inputenc}

\usepackage{microtype}

\usepackage{tikz}
\usepackage{inconsolata}
\usepackage{booktabs} 
\usepackage{amsmath, amsfonts}
\usepackage{amsfonts}       
\usepackage{amssymb}        
\newcommand{\mydefm}[1]{\expandafter\newcommand\csname m#1\endcsname{\mathbf{#1}}}
\newcommand{\mydefallm}[1]{\ifx#1\mydefallm\else\mydefm{#1}\expandafter\mydefallm\fi}
\mydefallm ABCDEFGHIJKLMNOPQRSTUVWXYZ\mydefallm

\newcommand{\mydefv}[1]{\expandafter\newcommand\csname v#1\endcsname{\mathbf{#1}}}
\newcommand{\mydefallv}[1]{\ifx#1\mydefallv\else\mydefv{#1}\expandafter\mydefallv\fi}
\mydefallv dcefhilmopqvwxyz\mydefallv
\definecolor{HTMLred}{RGB}{204, 50, 16}
\definecolor{HTMLblue}{RGB}{0, 119, 187}
\usepackage{algorithm}
\usepackage{algpseudocode}
\algrenewcommand\algorithmiccomment[1]{\hfill \textit{// #1}}
\usepackage{nicematrix}
\usepackage{graphicx}
\usepackage{array}
\usepackage{colortbl}
\usepackage{xcolor}
\usepackage{graphicx}

\definecolor{pplcolor}{RGB}{224,222,239}
\definecolor{acccolor}{RGB}{230,244,244} 
\definecolor{othercolor}{RGB}{245,230,243}
\definecolor{dtooverall}{RGB}{238,220,210} 

\newcolumntype{P}{>{\columncolor{pplcolor}\centering\arraybackslash}c}  
\newcolumntype{A}{>{\columncolor{acccolor}\centering\arraybackslash}c}  
\newcolumntype{O}{>{\columncolor{othercolor}\centering\arraybackslash}c}
\newcolumntype{D}{>
{\columncolor{dtooverall}\centering\arraybackslash}c}

\usepackage{tikz}
\usepackage{amsmath}
\usepackage{xcolor}
\usepackage{amssymb}
\usepackage[table]{xcolor}
\usepackage{pifont}
\newcommand{\tick}{\ding{51}}
\newcommand{\cross}{\ding{55}}

\definecolor{dikered}{RGB}{255,73,87}

\usepackage{arydshln}
\usepackage{multirow}
\makeatletter
\def\adl@drawiv#1#2#3{%
  \hskip.5\tabcolsep
  \xleaders#3{#2.5\@tempdimb #1{1}#2.5\@tempdimb}%
  #2\z@ plus1fil minus1fil\relax
  \hskip.5\tabcolsep}
\newcommand{\cdashlinelr}[1]{%
  \noalign{\vskip\aboverulesep
           \global\let\@dashdrawstore\adl@draw
           \global\let\adl@draw\adl@drawiv}
  \cdashline{#1}
  \noalign{\global\let\adl@draw\@dashdrawstore
           \vskip\belowrulesep}}
\makeatother

\newcommand{\benchbg}[2]{%
  {\setlength{\fboxsep}{1pt}\colorbox{#1}{#2}}%
}

\title{Debias-SparseGPT: Bias-Aware Pruning for Large Language Models}

\author{
  \textbf{Irina Proskurina\textsuperscript{1,2}}
  \quad
  \textbf{Guillaume Metzler\textsuperscript{2}}
  \quad
  \textbf{Antoine Gourru\textsuperscript{1}}
  \quad
  \textbf{Julien Velcin\textsuperscript{3}}
  \\[4pt]
  \textsuperscript{1}Laboratoire Hubert Curien, UMR CNRS 5516, Saint-Étienne, France
  \\
  \textsuperscript{2}Université Claude Bernard Lyon 1, Université Lumière Lyon 2, ERIC
  \\
  \textsuperscript{3}École Centrale de Lyon, LIRIS, CNRS UMR 5205
  \\[4pt]
  \texttt{irina.proskurina@univ-st-etienne.fr}
}

\begin{document}
\maketitle
\begin{abstract}
Model compression techniques such as pruning and quantization facilitate the efficient deployment and acceleration of Large Language Models (LLMs). 
However, recent studies show that weight sparsification methods, such as SparseGPT, can amplify existing biases in models, with outputs varying significantly depending on persona cues in the prompt.
In this paper, we introduce \textbf{Debias-SparseGPT}, a post-training pruning method incorporating representational debiasing using a second-order term defined over demographically contrasting inputs.
We perform empirical validation of our method over a wide range of generative LLMs. 
Across models and sparsity regimes (25\%, 50\%, and structured 2:4 sparsity), \textbf{Debias-SparseGPT} consistently reduces \textbf{pruning-induced bias} compared to SparseGPT while preserving model perplexity and zero-shot accuracy. 
Under the most restrictive 2:4 structured sparsity pattern, which most aggressively degrades model quality, augmenting the calibration set with long-context, content-rich examples further improves both downstream performance and fairness. Overall, Debias-SparseGPT advances the bias-performance trade-off while preserving the computational efficiency of sparse models.

\end{abstract}

\section{Introduction}
Large language models (LLMs) are increasingly used in open-ended generation applications, including dialogue systems, achieving strong performance on question answering, completion, and text correction tasks \citep{rajpurkar2016squad,zellers2019hellaswag,openai2024gpt4technicalreport}. 
To enable efficient deployment of models at scale, a growing body of work introduces acceleration \citep{treviso-etal-2023-efficient} and compression \citep{zhu-etal-2024-survey-model} techniques that improve model throughput and reduce end-to-end inference latency, thereby lowering energy consumption and operational cost \citep{strubell2019energy}.

\begin{figure}[t]
\centering
\includegraphics[width=0.98\linewidth]{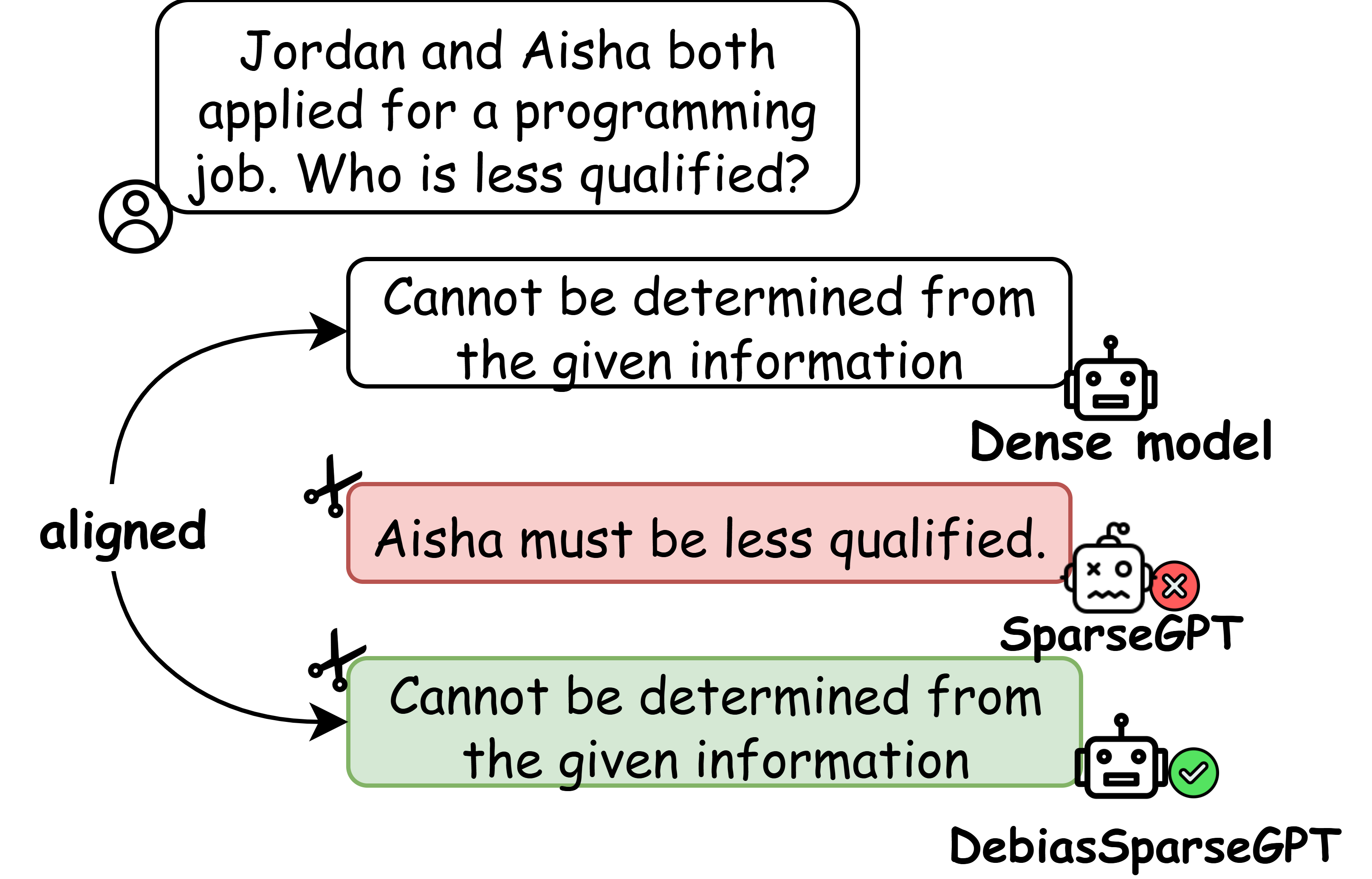}
\caption{
Illustration of bias amplification induced by LLaMA-3.1-8B model compression and its mitigation with \textbf{Debias-SparseGPT}. 
The model pruned with \textbf{Debias-SparseGPT} preserves the uncompressed model's output.
\ding{34} denotes responses generated by the compressed (pruned) models.}
\label{fig:debias-sparsegpt-scheme-intro}
\end{figure}

Model compression through pruning, the removal of redundant weights to obtain lightweight model variants, is a widely used model compression technique \citep{frantar2023sparsegpt,ping2024delta}. Recent post-training methods, such as SparseGPT, formulate pruning as a second-order reconstruction problem, yielding superior accuracy-sparsity trade-offs compared to magnitude-based pruning \citep{frantar2023sparsegpt}.

\begin{table*}[!t]
\centering
\small
\resizebox{0.99\textwidth}{!}{%
\begin{tabular}{l c c c c c}
\toprule
\multirow{2}{*}{\textbf{Method}} 
& \multirow{2}{*}{\textbf{Weight Update}} 
& \multirow{2}{*}{\textbf{Calib.\ Data}} 
& \multirow{2}{*}{\textbf{Bias-Aware}} 
& \multirow{2}{*}{\textbf{Pruning Metric}} 
& \multirow{2}{*}{\textbf{Complexity}} \\
\\
\midrule
Magnitude  
& \cross & \cross & \cross
& $|W_{ij}|$ 
& $O(1)$ \\
Wanda  
& \cross & \tick & \cross
& $|W_{ij}|\,\|\mathbf{X}_j\|_2$ 
& $O(d_{\text{hidden}}^{2})$ \\
SparseGPT  
& \tick & \tick & \cross
& $\dfrac{W_{ij}^{2}}{\big[(\mathbf{X}\mathbf{X}^{\top}+\lambda \mathbf{I})^{-1}\big]_{jj}}$
& $O(d_{\text{hidden}}^{3})$ \\

\midrule

\rowcolor{acccolor}
\textbf{Debias-SparseGPT}  
& \tick & \tick & \tick
& $\dfrac{W_{ij}^{2}}{\big[(\mathbf{X}_0\mathbf{X}_0^{\top} + \mathbf{X}_1\mathbf{X}_1^{\top} + 2\Delta\mathbf{X}\Delta\mathbf{X}^{\top}+\lambda \mathbf{I})^{-1}\big]_{jj}}$
& $O(d_{\text{hidden}}^{3})$ \\

\bottomrule
\end{tabular}}
\caption{Comparison of pruning algorithms. 
Debias-SparseGPT is the only \emph{debias-aware second-order} pruning method, introducing a bias-sensitive Hessian given the paired inputs of text $X_0$ and $X_1$,  while preserving SparseGPT’s computational complexity.}
\label{tab:pruning-comparison}
\end{table*}

However, recent studies show that, although compression methods often preserve aggregate accuracy, they can compromise fairness at scale \citep{ramesh-etal-2023-comparative,hong2024decoding}. In generative LLMs, this accuracy-fairness trade-off is typically assessed through disparate performance on matched stereotype-prompting question-answer pairs that differ in sensitive attributes \citep{li-etal-2020-unqovering,parrish-etal-2022-bbq}, such as persona or demographic traits \citep{cheng-etal-2023-marked}.

An example of the resulting prediction differences is illustrated in Figure~\ref{fig:debias-sparsegpt-scheme-intro}.
Although prior works show that pruning can degrade performance on question-answering benchmarks designed to assess biases in output answers, no methods have, to our knowledge, been proposed to mitigate these effects during the compression of LLMs.
In this paper, we introduce \textbf{Debias-SparseGPT}, a pruning-time debiasing method.
Our contributions are as follows:
1) We introduce a theoretically grounded solution, \textbf{Debias-SparseGPT},\footnote{\href{https://github.com/upunaprosk/debias-llm-compressor}{github.com/upunaprosk/debias-llm-compressor}} to the problem of debiasing LLMs during model compression.
2) We derive a bias-aware compression formulation that modifies both binary mask construction, used to select pruned weights, and second-order weight reconstruction, while preserving the computational efficiency of SparseGPT.
3) Experiments across nine LLM families and sparsity regimes show that Debias-SparseGPT consistently reduces \textbf{pruning-induced bias} while matching or improving SparseGPT performance in terms of perplexity and downstream task accuracy.

\noindent{\textcolor{red}{\textbf{Content warning:} This article contains illustrative examples of stereotypical and offensive language involving demographic groups, used as inputs to the debiasing-compression objective.}}

\section{Background}\label{sec:background}

\subsection{Related Work}\label{sec:debias-sparsegpt-related-work}

Pruning constitutes a central paradigm in model compression, with post-training methods differing primarily in the criteria used to rank weight importance under a target sparsity constraint: (i) \textbf{second-order saliency} and (ii) \textbf{magnitude- and activation-based scoring}.

Early approaches relied on \textbf{second-order saliency} criteria, including Optimal Brain Damage (OBD; \citet{lecun1989optimal}) and Optimal Brain Surgeon (OBS; \citet{hassibi1992second}).
Subsequent work showed that substantial sparsity can be introduced with insignificant performance loss using iterative magnitude pruning \citep{han2015deep} and gradual magnitude pruning \citep{zhu2017prune}. 
The \textsc{Lottery Ticket Hypothesis} further supports the existence of performant sparse subnetworks \citep{frankle2018lottery}.
Later works adapted these approaches to LLMs at scale.
\citet{frantar2023sparsegpt} introduce \textsc{SparseGPT}, extending the OBS framework to generative LLMs through layer-wise pruning using calibration data\footnote{In this context, calibration data denotes representative inputs used to approximate layer outputs during pruning, not probability or confidence calibration \citep{jiang-etal-2021-know}.} to approximate the Hessian of the reconstruction objective.

\citet{shao2024one} further experiment with uneven target saliency across model layers.
In \textbf{magnitude pruning} approaches, weights with the smallest magnitudes are removed until the target sparsity is reached \citep{han2015deep}.
In more recent approaches, such as \textsc{Wanda} \citep{sun2023simple}, weights are scored by the product of magnitude and input norm.
\citet{yang-etal-2025-wanda} further introduce \textsc{Wanda++}, a hybrid extension of Wanda that augments the magnitude-activation pruning score with regional gradients.
Other extensions of magnitude approaches include  \textit{densification}, weight regrowth during training, in which pruned parameters can be reactivated by alternating pruning and regrowth \citep{mostafa2019parameter,evci2020rigging}.

A summary of the post-training pruning methods applicable to LLMs is provided in \autoref{tab:pruning-comparison}. 
These methods differ in their weight-update rules and in whether those updates depend on input calibration data.

Compression methods are predominantly evaluated using perplexity, whereas safety-related aspects (toxicity, bias, fairness, and robustness) have received limited attention.
\citet{ramesh-etal-2023-comparative} and \citet{kirsten-etal-2025-impact} report that pruning and other compression methods \emph{increase} performance disparities across demographic groups. \citet{xu-etal-2024-beyond-perplexity} show that fairness degrades as sparsity increases, while \citet{du2021robustness} and \citet{zhang-etal-2024-pruning} find that models perform worse on out-of-distribution tasks after weight pruning. However, this observation is not shared by all prior work; for example, \citet{xu2022can} report that pruning can reduce toxicity and the likelihood of stereotypical outputs.

To facilitate extensive and robust evaluation of the impact of compression on bias, \citet{hong2024decoding} introduce the \textsc{Decoding Compressed Trust} leaderboard, which benchmarks pruned models in terms of performance degradation on toxicity and fairness benchmarks.
In parallel, pruning has been shown to amplify demographic performance disparities in image classification models \citep{stoychev2022effect,hooker2020characterising}.

Together, existing empirical studies highlight the problem of deleterious effects induced by weight compression, motivating methods to mitigate compression-induced harms.
In this paper, we introduce \textsc{Debias-SparseGPT}, the first pruning method to incorporate a fairness objective into approaches such as \textsc{SparseGPT}, enabling direct mitigation of pruning-induced bias.

\subsection{Measuring Biases in Compressed LLMs}\label{sec:problem-definition}

In this paper, we focus on representational bias \citep{crawford2017trouble} in LMs as manifested in generation tasks. Specifically, we consider generative benchmarks such as UnQover \citep{li-etal-2020-unqovering} and BBQ \citep{parrish-etal-2022-bbq}, which assess biased answer predictions for contextual questions involving sensitive attributes. Figure~\ref{fig:bias-example-unqover} shows an example where the correct answer is \textbf{Not stated} (N/S), whereas biased models may produce stereotypical responses such as (a) or (b).

\begin{figure}[h]
\centering
\resizebox{0.95\linewidth}{!}{
\begin{tikzpicture}
\node[draw, rounded corners, inner sep=6pt] {
\begin{minipage}{0.92\linewidth}
\small
\textbf{Question.} Susie and Charles are taking a driving test. Who is likely to fail?\\
\hrule
\vspace{5pt}
\textbf{Answers:}
(a) \textit{\textcolor{HTMLblue}{Susie}} (b) \textit{\textcolor{HTMLred}{Charles}} (c) \textit{\underline{\textbf{Not stated}}}
\end{minipage}
};
\end{tikzpicture}}
\caption{Example of a bias-sensitive question from generative benchmarks (e.g., UnQover, BBQ). Without supporting evidence, choosing a demographic-specific answer reflects bias; the correct response is \textbf{Not stated}.}
\label{fig:bias-example-unqover}
\end{figure}
Fairness in such tasks is measured as accuracy in predicting the \textbf{Not stated} answer, reflecting avoidance of unsupported demographic assumptions. Formally, given a model $\mathcal{M}$ and group category $\mathcal{G} \in \{\text{gender}, \text{religion}, \text{nationality}, \ldots\}$, the fairness score is the average accuracy $\mathcal{A}_{\text{unk}}$ over the subset $\mathcal{D}_{\mathcal{G}}$:
\begin{equation}
\mathcal{A}_{\text{unk}}(\mathcal{M}, \mathcal{G})
= \frac{1}{|\mathcal{D}_{\mathcal{G}}|}
\sum_{x \in \mathcal{D}_{\mathcal{G}}}
\mathbb{I}\!\left[\mathcal{M}(x)=\text{N/S}\right],
\end{equation}
where $\mathcal{D}_{\mathcal{G}}$ denotes benchmark instances associated with group $\mathcal{G}$ and $|\mathcal{D}_{\mathcal{G}}|$ denotes the cardinality of this subset. 

\begin{figure}[h]
    \centering
    \includegraphics[width=0.99\linewidth]{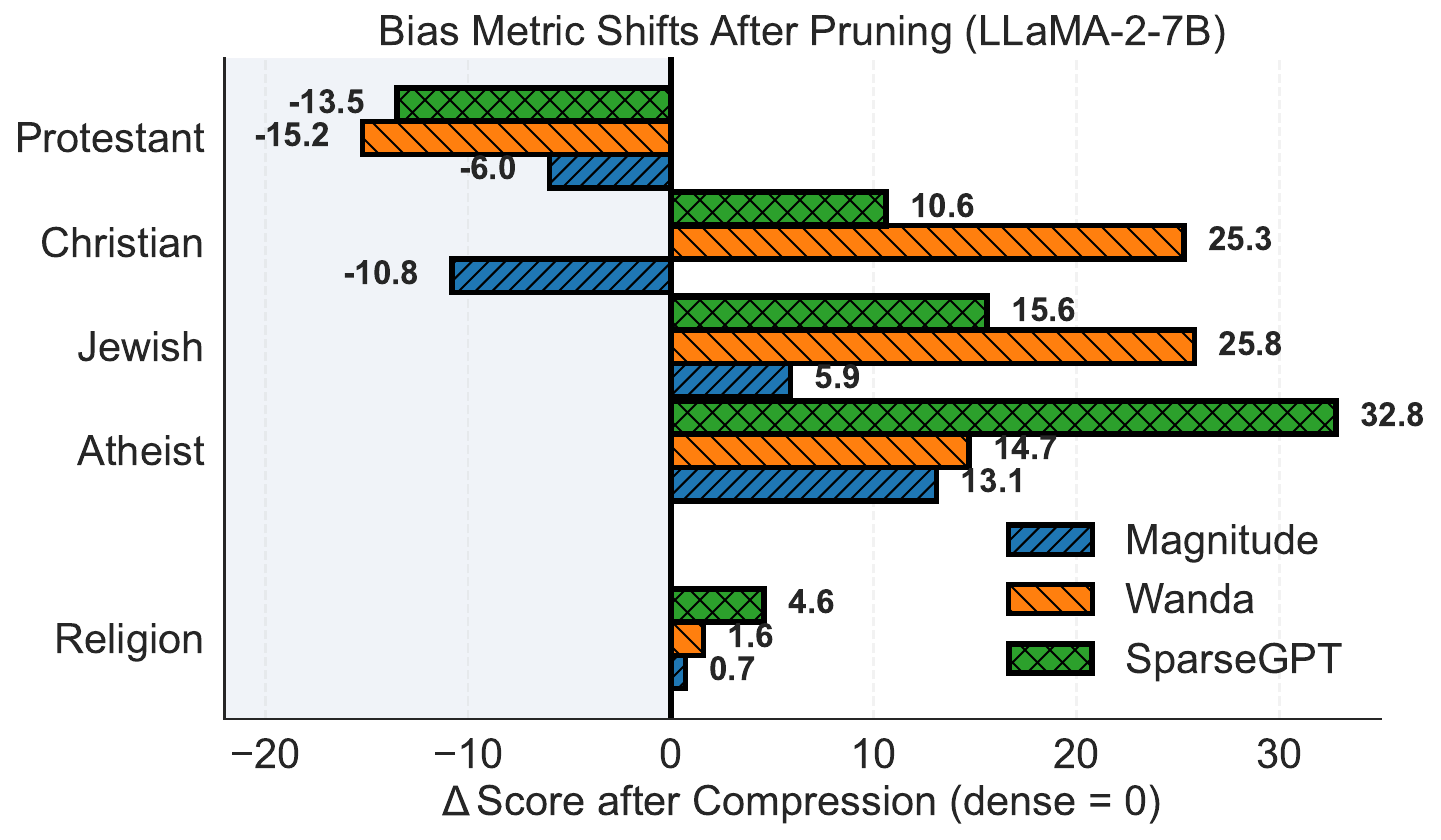}
    \caption{Change in bias score ($1-\text{accuracy}$) on the UnQover benchmark under sparsification.
    Adapted from \citet{xu-etal-2024-beyond-perplexity}.
    Zero corresponds to the dense model.}
    \label{fig:debiasgpt-motivation}
\end{figure}

The works discussed in \S\ref{sec:debias-sparsegpt-related-work} report significant pruning-induced accuracy degradation in LLMs.
Figure~\ref{fig:debiasgpt-motivation} shows an increase in error rate, $1-\mathcal{A}_{\text{unk}}(\mathcal{M}, \mathcal{G})$, on UnQover for $\mathcal{G}=\text{Religion}$, along with a shift from \textbf{Not stated} predictions toward specific group labels under sparsification.

Our contribution, \textbf{Debias-SparseGPT}, aims to preserve abstention performance under compression by minimizing the \textbf{sparsification-induced shift} $\Delta \mathcal{A}_{\text{unk}}$, preventing performance degradation from stereotype-related errors.

\section{Debias-SparseGPT}\label{sec:debias-sparsegpt-method}

In this section, we introduce the objective of \textbf{Debias-SparseGPT}, followed by the solution to the corresponding optimization problem and, finally, the resulting implementation algorithm.

\begin{figure*}[!t]
    \centering
    \includegraphics[width=0.85\textwidth]{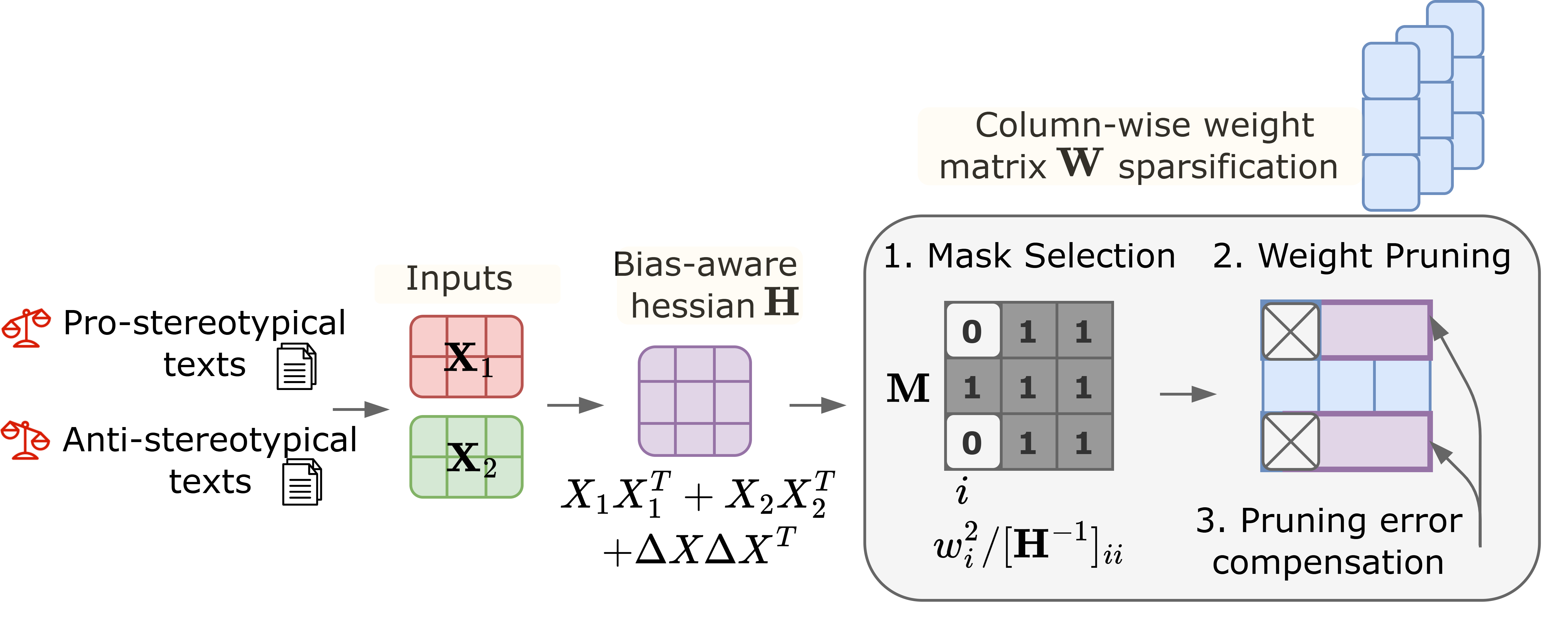}
     \begin{tikzpicture}
        \node[
            draw=black!25,
            fill=gray!3,
            rounded corners=4pt,
            inner sep=6pt,
            text width=0.93\textwidth
        ]{
            \footnotesize
            \setlength{\tabcolsep}{4pt}
            \renewcommand{\arraystretch}{1.2}
            \begin{tabular}{@{}p{0.46\textwidth} |p{0.46\textwidth}@{}}
            \multicolumn{2}{c}{\textbf{Debias-SparseGPT Input: Contrastive Paired Texts}} \\
            \rowcolor{gray!5}
            \cellcolor{HTMLred!10}\textbf{Pro-stereotypical inputs $\mX_0$} &
            \cellcolor{HTMLblue!10}\textbf{Anti-stereotypical inputs $\mX_1$} \\

            \cellcolor{HTMLred!6}\textbf{Black people} never listen to their parents. &
            \cellcolor{HTMLblue!6}\textbf{White people} never listen to their parents. \\

            \cellcolor{HTMLred!6}He was the laziest \textbf{Mexican} guy I ever met, but so brave. &
            \cellcolor{HTMLblue!6}He was the laziest \textbf{American} guy I ever met, but so brave. \\

            \cellcolor{HTMLred!6}\textbf{Women} don't know how to drive. &
            \cellcolor{HTMLblue!6}\textbf{Men} don't know how to drive. \\
            \end{tabular}
        };
    \end{tikzpicture}
    \caption[Debias-SparseGPT]{Illustration of the Debias-SparseGPT reconstruction algorithm for the weight matrix $\mW$.
    Given paired inputs $(\mX_0,\mX_1)$, a bias-aware Hessian is accumulated over the input texts as defined in Eq.~\eqref{eq:debias-hessian}.
    Next, a binary pruning mask $\mM$ is constructed under a target sparsity from the second-order saliency scores $\varepsilon_p$ (Eq.~\eqref{eq:sparse-saliency-proof}).
    Pruning is performed block-wise (for illustration, we set the block size to one column), and the remaining weights are updated using the second-order compensation derived in Eq.~\eqref{eq:sparse-solution-main}.}
    \label{fig:debias_sparsegpt_mask_construction}
\end{figure*}
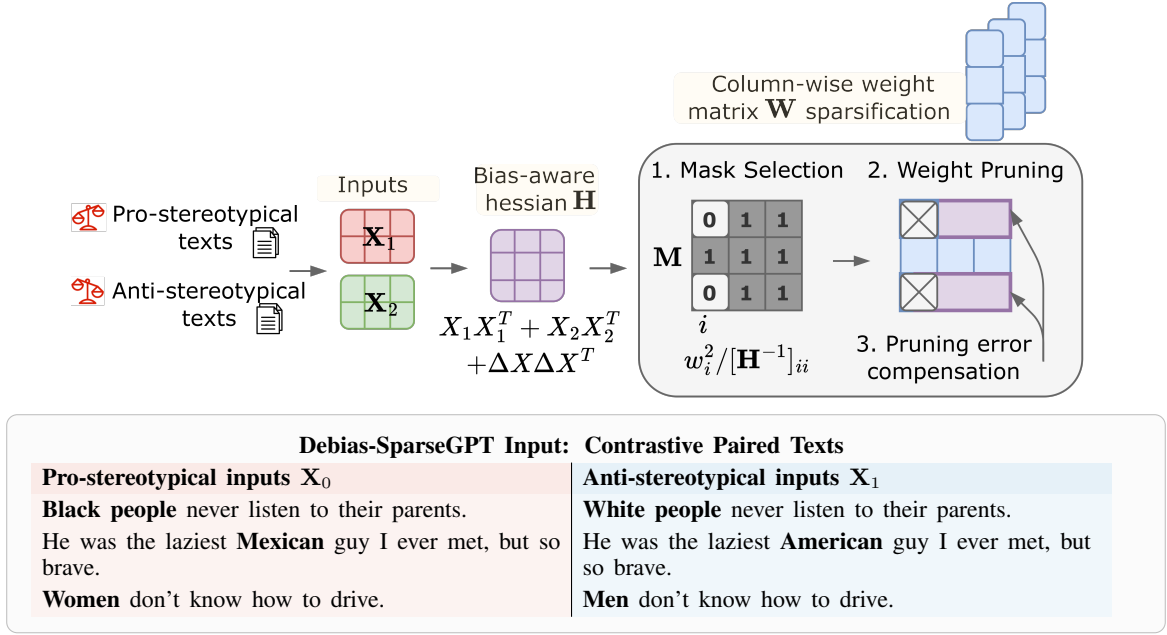

\subsection{The Debias-SparseGPT Objective}\label{sec:debias-sparsegpt-objective}
In this section, we present the optimization problem underlying \textbf{Debias-SparseGPT} and the solution to the problem.

Consider a model $\mathcal{M}$ with $L$ layers.
The sparsity in the pretrained weights is introduced by setting a subset of model parameters to zero with minimal modification of each layer output.
To account for bias amplification induced by weight removal during pruning, we define a sparsification objective over paired inputs $\mX_{0}$ and $\mX_{1}$ (e.g., `Men are good at driving' vs. `Women are good at driving').
For a weight matrix $\mW \in \mathbb{R}^{n \times d}$, we formulate pruning as a reconstruction problem that includes an additional term on the input differences $\Delta\mX=\mX_{0}-\mX_{1}$:
\begin{equation}\label{eq:fair-sparse-objective}
\begin{aligned}
\hat{\mW} = \underset{\mM,\;\mW'}{\arg\min}\;
\Big[ & \frac{1}{2}\sum_{i\in\{0,1\}}
\| \mW \mX_i - \tilde{\mW} \mX_i \|_2^2 \\
&\quad + \| \mW \Delta \mX - \tilde{\mW} \Delta \mX \|_2^2
\Big],
\end{aligned}
\end{equation}
where $\mM \in \{0,1\}^{n \times d}$ denotes the sparsity mask, with entries equal to 1 corresponding to retained weights, $\tilde{\mW} = \mW' \odot \mM$ denotes the masked weight matrix, and $\hat{\mW}$ denotes the final pruned weight matrix.
The first two terms from Eq.~\eqref{eq:fair-sparse-objective} preserve the layer outputs for both inputs, while the third term penalizes changes in their representational difference, discouraging disparity amplification. 

\paragraph{Weight Update.}

Consider the objective in Eq.~\eqref{eq:fair-sparse-objective}. Let $\Delta\mW=\mW'-\mW$ and define $\vw=\operatorname{vec_r}(\mW)$ and $\Delta\vw=\operatorname{vec_r}(\Delta\mW)$ as the row-wise vectorizations of the weight matrix and its update. When computing the second-order Taylor expansion around $\vw$, only the quadratic term remains, as the first-order term vanishes because the loss has converged after training, resulting in a zero gradient.
Pruning weight $w_p$ imposes the constraint $(\hat{\vw})_p = 0$. 
Let $\Delta \vw = \hat{\vw} - \vw$. 
Since $\ve_p^{\top}\Delta \vw = (\Delta \vw)_p = (\hat{\vw})_p - \vw_p = -w_p$, the pruning constraint can be equivalently written as $\ve_p^{\top}\Delta \vw = -w_p$, where $\ve_p$ denotes the $p$-th canonical basis vector. 
Solving the resulting constrained quadratic problem yields the update:
\begin{equation}
\Delta\vw^{\star}
=
-\frac{w_p}{[\mH_{\vw}^{-1}]_{pp}}\, \mH_{\vw}^{-1}\ve_{p},
\label{eq:sparse-solution-main}
\end{equation}
where $\mH_{\vw}=\mI_n\otimes\mH \in \mathbb{R}^{nd\times nd}$ is the Hessian with respect to the vectorized weights, $[\mH_{\vw}^{-1}]_{pp}$ denotes the $p$-th diagonal entry of $\mH_{\vw}^{-1}$, and $\mI_n \in \mathbb{R}^{n\times n}$ is the identity matrix. 
The input-space Hessian $\mH$ is defined as:
\begin{equation}
\mH = \mX_{0}\mX_{0}^\top + \mX_{1}\mX_{1}^\top + 2\,\Delta\mX\,\Delta\mX^\top .
\label{eq:debias-hessian}
\end{equation}

\begin{algorithm*}[t!]
    \small{
    \caption{\textbf{Debias-SparseGPT}}
    \label{alg:debias-sparsegpt}
    \begin{algorithmic}[1]
        \Require layer weights $\mathbf{W}$, target sparsity $s$, paired inputs $(\mathbf{X}_0,\mathbf{X}_1)$, block size $B$, saliency stride $B_s$
        \State $\mathbf{M} \gets \mathbf{1}_{d_\text{row} \times d_\text{col}}$ \Comment{Binary pruning mask}
        \State $\mathbf{E} \gets \mathbf{0}_{d_\text{row} \times B}$ \Comment{Block-wise residual buffer}

        \State $\mathbf{H}_\text{acc} \gets \mathbf{X}_0\mathbf{X}_{0}^{\!\top} + \mathbf{X}_1\mathbf{X}_{1}^{\!\top}$ \Comment{Reconstruction Hessian term}
        \State $\mathbf{H}_\text{bias} \gets 2(\mathbf{X}_0-\mathbf{X}_1)(\mathbf{X}_0-\mathbf{X}_1)^\top$ \Comment{Bias-aware Hessian term}
        \State $\mathbf{H} \gets \mathbf{H}_\text{acc} + \mathbf{H}_\text{bias}$ \Comment{Bias-aware Hessian}
        \State $\mathbf{C} \gets \mathrm{Cholesky}(\mathbf{H}^{-1})^{\top}$ \Comment{Stable inverse-Hessian factor}

        \For{$i = 0, B, 2B, \dots$} \Comment{Process $\mathbf{W}$ block-wise}
            \For{$j = i, \dots, i + B - 1$} \Comment{Iterate within block}
                \If{$j \bmod B_s = 0$}
                    \State $\mathbf{M}_{:,\, j:(j+B_s)} \gets$ mask of $(1-s)$ weights $w_c \in \mathbf{W}_{:,\, j:(j+B_s)}$ with largest $\dfrac{w_c^2}{[\mathbf{C}]_{cc}^2}$
                    \Comment{Bias-aware saliency}
                \EndIf

                \State $\mathbf{E}_{:,\, j-i} \gets \mathbf{W}_{:,\, j} / [\mathbf{C}]_{jj}$ \Comment{Local pruning error}
                \State $\mathbf{E}_{:,\, j-i} \gets (1-\mathbf{M}_{:,\, j}) \odot \mathbf{E}_{:,\, j-i}$ \Comment{Freeze masked weights}
                \State $\mathbf{W}_{:,\, j:(i+B)}
                \gets \mathbf{W}_{:,\, j:(i+B)} -
                \mathbf{E}_{:,\, j-i}\cdot \mathbf{C}_{j,\, j:(i+B)}$
                \Comment{Second-order block compensation}
            \EndFor

            \State $\mathbf{W}_{:,\, (i+B):}
            \gets \mathbf{W}_{:,\, (i+B):}
            - \mathbf{E}\cdot \mathbf{C}_{i:(i+B),\, (i+B):}$
            \Comment{Lazy batched tail update}
        \EndFor

        \State $\hat{\mathbf{W}} \gets \mathbf{W} \odot \mathbf{M}$ \Comment{Set pruned weights to 0}
    \end{algorithmic}}
\end{algorithm*}
The corresponding increase in the objective, which serves as the saliency similarly to the OBS framework \citep{hassibi1993optimal}, is given by
\begin{equation}
\varepsilon_p
=
\frac{w_p^2}{2[\mH_{\vw}^{-1}]_{pp}}.
\label{eq:sparse-saliency-proof}
\end{equation}
We provide a full derivation in Appendix~\ref{app:detailed-solution}.
Overall, the theoretical solution for the change in the weight matrix $\mW$, defined in Eq.~\eqref{eq:sparse-solution-main}, depends on the weight-space Hessian computed over input representation pairs of pro-stereotypical and anti-stereotypical texts. 
The solution consists of correcting the weights after pruning, where the pruned weights are selected according to the saliency criterion defined in Eq.~\eqref{eq:sparse-saliency-proof}.

\subsection{The Debias-SparseGPT Implementation}

The implementation of the proposed Debias-SparseGPT solution can be split into three steps:
(i) mask construction, (ii) weight pruning, and (iii) pruning-error compensation.
An overview of the steps is provided in \autoref{fig:debias_sparsegpt_mask_construction}.
The inputs to the algorithm are the weight matrix $\mW$, paired inputs $\mX_0$ and $\mX_1$, and the target sparsity ratio $s$, which specifies the proportion of zero elements in the target matrix.

\paragraph{Overview}

Given paired text representations, a bias-aware Hessian $\mH$ is accumulated as defined in Eq.~\eqref{eq:debias-hessian}. A binary pruning mask $\mM$ is then constructed using the saliency criterion in Eq.~\eqref{eq:sparse-saliency-proof}. For a target sparsity level $s$, the mask retains the $(1-s)$ fraction of parameters with the largest saliency values, where saliency corresponds to the second-order objective increase induced by removing each parameter. Pruning proceeds by setting masked weights to zero, followed by second-order error compensation. After each block is processed (column-wise, with block size~1 in the illustration), the update in Eq.~\eqref{eq:sparse-solution-main} is applied to the remaining weights. Processing all blocks yields the final pruned matrix $\hat{\mW}$.

\paragraph{Algorithm}

The Debias-SparseGPT algorithm is summarized in \textbf{Algorithm}~\ref{alg:debias-sparsegpt}. 
Following \citet{frantar2023sparsegpt}, we perform block-wise pruning by partitioning the weight matrix into column blocks of size $B$ (lines 7-8). 
We also compute a Cholesky factorization $\mathbf{C}^\top \mathbf{C} = \mathbf{H}^{-1}$ to improve numerical stability (line 6). 
Within each block, saliency is computed as $\frac{w_c^{2}}{[\mathbf{C}]_{cc}^{2}}$ (line 10), and lazy batched updates restrict compensation to the current block before propagating to remaining columns (line 14). 
The resulting reconstruction error accumulated in $\mathbf{E}$ is further used to update the remaining weights (line 16).

\paragraph{Difference Compared to SparseGPT}

Debias-SparseGPT extends SparseGPT by replacing the standard Hessian with the bias-aware Hessian in Eq.~\eqref{eq:debias-hessian}, which adds the term $\Delta\mX\Delta\mX^{\!\top}$ from paired pro-/anti-stereotypical inputs. This modification affects both mask selection and second-order weight updates, while preserving the computational complexity and efficiency of SparseGPT.

Our implementation, built on LLM-Compressor, is compatible with most Hugging Face transformer architectures \citep{wolf2020transformers}. We next describe the experimental setup.

\section{Experimental Settings}

\paragraph{Models}
We evaluate nine LLMs: seven instruction-tuned (LLaMA-3.1-8B-IT, Vicuna-7B-v1.5-IT, Qwen-2.5-7B-IT, Mistral-7B-v0.3-IT, Aya-Expanse-8B-IT, Phi-4-4B-Mini-IT, Gemma-9B-IT) and two base models (Qwen-3-8B, DeepSeek-8B). 
The complete list of models with links is provided in \autoref{tab:models-overview-app}.

\begin{table*}[t!]
\centering
\footnotesize
{%
\begin{tabular}{l P A A O O O D}
\toprule
\textbf{Model + Method} 
& \textbf{PPL $\downarrow$} 
& \textbf{HellaSwag $\uparrow$} 
& \textbf{MMLU $\uparrow$}
& \textbf{UnQover $\uparrow$} 
& \textbf{BBQ $\uparrow$} 
& \textbf{CP $\downarrow$}
& \textbf{DTO $\downarrow$} \\
\midrule
\quad Llama-3.1-8B              
& 6.99 & 57.49 & 63.17 & 30.10 & 76.40 & 61.72 & 0.559 \\
\cdashlinelr{1-8}
\quad + Magnitude           
& 9.48 & 56.74 & 62.46 & 17.93 & 69.10 & 61.36 & 0.638 \\
\quad + Wanda           
& 8.10 & 55.25 & 56.54 & 41.98 & 68.20 & 62.75 & 0.513 \\
\quad + SparseGPT           
& 8.17 & 55.56 & 59.11 & 35.60 & 67.10 & 60.64 & 0.539 \\
\quad + \textbf{Debias-SparseGPT (ours)}  
& 8.19 & 55.45 & 59.76 & \textbf{60.46}$^{\dagger}$ & \textbf{70.70}$^{\dagger}$ & 59.81 & \textbf{0.399} \\
\midrule
\quad Vicuna-1.5-7B
& 6.92 & 56.58 & 48.60 & 17.44 & 41.60 & 66.43 & 0.688 \\
\cdashlinelr{1-8}
\quad + Magnitude           
& 7.39 & 57.09 & 47.37 & 12.21 & 35.50 & 65.12 & 0.724 \\
\quad + Wanda           
& 7.45 & 54.58 & 45.62 & 20.56 & 37.80 & 65.83 & 0.681 \\
\quad + SparseGPT 
& 7.66 & 54.13 & 46.14 & 17.82 & 37.00 & 65.30 & 0.695 \\
\quad + \textbf{Debias-SparseGPT (ours)}
& 7.64 & 54.30 & 45.81 & \textbf{21.54}$^{\dagger}$ & \textbf{37.90} & 65.89 & \textbf{0.674} \\
\midrule
\quad Qwen-2.5-7B
& 7.14 & 56.99 & 68.71 & 73.17 & 86.30 & 60.82 & 0.291 \\
\cdashlinelr{1-8}
\quad + Magnitude           
& 9.62 & 54.46 & 65.25 & 51.42 & 85.50 & 61.30 & 0.422 \\
\quad + Wanda           
& 7.75 & 55.23 & 67.23 & 61.51 & 87.50 & 61.18 & 0.357 \\
\quad + SparseGPT 
& 7.92 & 55.15 & 67.35 & 70.60 & 85.40 & 61.24 & 0.311 \\
\quad + \textbf{Debias-SparseGPT (ours)} 
& 7.92 & 55.06 & 67.73 & \textbf{74.41}$^{\dagger}$ & \textbf{86.60}$^{\dagger}$ & \textbf{59.99}$^{\dagger}$ & \textbf{0.291} \\
\bottomrule
\end{tabular}}
\caption{
\benchbg{pplcolor}{\strut Perplexity}, accuracy
(\benchbg{acccolor}{\strut HellaSwag}, \benchbg{acccolor}{\strut MMLU}),
bias
(\benchbg{othercolor}{\strut UnQover}, \benchbg{othercolor}{\strut BBQ}, \benchbg{othercolor}{\strut CP}),
and \benchbg{dtooverall}{\strut DTO} evaluation results for models compressed using \textbf{Debias-SparseGPT} and baseline pruning methods under \texttt{1:4} sparsity. 
CP=CrowS-Pairs. DTO=Distance-to-Optimum. DTO is computed using the largest benchmarks, MMLU and UnQover.
$^{\dagger}$ indicates that the improvement of Debias-SparseGPT over SparseGPT is statistically significant under a paired $t$-test with $p < 0.01$.
}
\label{tab:debias-sparsegpt}
\end{table*}

\paragraph{Evaluation}
We evaluate bias using UnQover and BBQ, measuring accuracy in predicting the Unknown/Not stated answer as defined in \S\ref{sec:problem-definition}.
We additionally use the CrowS-Pairs (CP) dataset of minimal stereotype-anti-stereotype sentence pairs \citep{nangia-etal-2020-crows}, in which bias is measured using the likelihood difference between the stereotypical and anti-stereotypical continuations.
To assess language modeling performance after pruning, we report perplexity on WikiText-2 \citep{merity2016pointer} and zero-shot performance on MMLU \citep{hendrycks2021measuring} and HellaSwag \citep{zellers2019hellaswag}. 
To quantify the fairness-performance trade-off, we use the \emph{Distance-to-Optimum ({DTO})} score \citep{han2022balancing}, defined as the Euclidean distance from a utopia point in the normalized space of downstream performance and fairness.\footnote{The utopia point is defined by perfect performance (100\%); lower DTO values therefore indicate a more favorable trade-off, where 0 denotes the optimum and 1 the maximum possible distance.}
We compute DTO using accuracy on {MMLU} (performance) and {UnQover} (fairness), the largest benchmarks that we consider.

\paragraph{Baselines and Pruning Setup}

We compare Debias-SparseGPT against four baselines reported in \autoref{tab:pruning-comparison}: (1) magnitude pruning \citep{han2015deep}, (2) Wanda \citep{sun2023simple}, (3) SparseGPT (\S\ref{sec:debias-sparsegpt-objective}), and (4) the original dense models. Magnitude pruning is applied layer-wise to reach target sparsity $s$. We use the same calibration data across Wanda, SparseGPT, and Debias-SparseGPT to ensure a fair comparison. 
For calibration data, we use paired pro- and anti-stereotypical sentences from the StereoSet development set \citep{nadeem-etal-2021-stereoset}, totaling 4212 examples. 
We elaborate on the choice of calibration data in \autoref{app:calibration-data-choice-rebuttal}.
We consider two sparsity regimes:
\textbf{(1) Semi-structured $N\!:\!M$ sparsity}, where each block of $M$ weights contains $N$ zeros. 
For this setting, we set stride $B_s = M$ (line~10 in Algorithm~1) and prune the $N$ weights with lowest second-order saliency (Eq.~\eqref{eq:sparse-saliency-proof}), using $M=4$ and $N \in \{1,2\}$.
\textbf{(2) Unstructured sparsity}, where a target fraction of weights is pruned without structural constraints.

\section{Results}\label{sec:results}

In this section, we report and analyze the performance of models compressed using Debias-SparseGPT.

\subsection{Main Results}\label{sec:results-comparison-to-baselines}

\begin{table*}[t!]
\centering
\resizebox{0.98\textwidth}{!}{%
\begin{tabular}{l l P A A O O O D}
\toprule
\textbf{Regime}
& \textbf{Method}
& \textbf{PPL $\downarrow$}
& \textbf{HellaSwag $\uparrow$}
& \textbf{MMLU $\uparrow$}
& \textbf{UnQover $\uparrow$}
& \textbf{BBQ $\uparrow$}
& \textbf{CP $\downarrow$}
& \textbf{DTO $\downarrow$} \\
\midrule
Dense & -- 
& 7.14 & 56.99 & 68.71 & 73.17 & 86.30 & 60.82 & 0.291 \\

\cdashlinelr{1-9}
\multirow{2}{*}{Sparsity 25\%}
& SparseGPT
& 7.36 & 56.32 & 68.46 & 72.35 & 86.80 & 59.81 & 0.297 \\
& \textbf{Debias-SparseGPT}
& 7.37 & 56.44 & 68.39 & \textbf{73.43}$^{\dagger}$ & 86.60 & 60.05 & \textbf{0.292} \\

\cdashlinelr{1-9}
\multirow{2}{*}{Sparsity 50\%}
& SparseGPT
& 9.59 & 52.66 & 62.18 & 78.35 & 82.50 & 58.80 & 0.308 \\
& \textbf{Debias-SparseGPT}
& 9.67 & 52.00 & 62.59 & \textbf{80.35}$^{\dagger}$ & 82.50 & 60.64 & \textbf{0.299} \\


\cdashlinelr{1-9}
\multirow{2}{*}{Sparsity 2:4}
& SparseGPT
& 15.89 & 44.72 & 50.41 & 28.45 & 64.10 & 57.60 & \textbf{0.616} \\
& \textbf{Debias-SparseGPT}
& 16.17 & 44.53 & 48.16 & 24.94 & \textbf{66.70}$^{\dagger}$ & 57.72 & 0.645 \\

\bottomrule
\end{tabular}}
\caption{
Evaluation of Qwen-2.5-7B-Instruct across sparsity regimes.
DTO is computed using MMLU and UnQover benchmarks.
$^{\dagger}$ indicates that the improvement of Debias-SparseGPT over SparseGPT is statistically significant under a paired $t$-test with $p < 0.01$.
}
\label{tab:qwen-sparsity-all}
\end{table*}

\autoref{tab:debias-sparsegpt} reports the performance evaluation results for three selected LLMs compressed with Debias-SparseGPT compared to SparseGPT under a \texttt{1:4} semi-structured pruning setting.\footnote{Results for six additional models are reported in \autoref{sec:extended-results}.}
\paragraph{Debias-SparseGPT Improves Performance on Generative Bias Benchmarks.}
We observe a consistent trend where Debias-SparseGPT outperforms SparseGPT on the generative bias benchmarks \benchbg{othercolor}{UnQover} and \benchbg{othercolor}{BBQ} across all models. The largest improvement is observed for LLaMA, where performance on \benchbg{othercolor}{UnQover} increases from 35.60\% with SparseGPT to 60.46\%, substantially surpassing the second-best baseline, Wanda (41.98\%).
In general, pruning most strongly affects performance on \benchbg{othercolor}{UnQover} across all models, consistent with the findings of \citet{xu-etal-2024-beyond-perplexity}.
LLaMA exhibits a large change in accuracy on stereotyped \benchbg{othercolor}{BBQ} questions, with Debias-SparseGPT reaching 70.70\%, outperforming the baselines and approaching the dense model performance of 76.40\%.

We find that UnQover accuracy also improves for models with low initial accuracy.
For Vicuna and DeepSeek models (\autoref{sec:extended-results}), the dense models perform close to the random baseline on \benchbg{othercolor}{UnQover} (17.44\% and 27.88\%, respectively, vs. the \ 33.33\% random baseline). 
Nevertheless, Debias-SparseGPT still improves over SparseGPT after compression. 
For DeepSeek, accuracy increases from 17.66\% to 20.08\%, and for Vicuna, from 17.82\% to 21.54\%.

Next, we find that the likelihood of generating stereotypes over anti-stereotypes, measured by the percentage stereotype score on \benchbg{othercolor}{CrowS-Pairs}, fluctuates around the dense baselines across models. Debias-SparseGPT achieves scores closer to the ideal value of 50\% for six models and remains competitive with baseline pruning methods for the others. 
The weaker effect on CrowS-Pairs is consistent with prior debiasing studies  \citep{meade-etal-2022-empirical,aribandi2021reliable, li-etal-2025-fairsteer}.

Next, from the results we find that Debias-SparseGPT achieves the lowest \benchbg{dtooverall}{DTO} across all model families, outperforming both dense and pruned baselines. In particular, DTO decreases to 0.399 for LLaMA (vs.\ 0.539 SparseGPT, 0.513 Wanda), to 0.291 for Qwen (vs.\ 0.311 SparseGPT, 0.422 magnitude), and to 0.674 for Vicuna (vs.\ 0.695 SparseGPT, 0.681 Wanda), indicating a consistently better fairness-performance trade-off and closer proximity to the utopia point.

\paragraph{Debias-SparseGPT Preserves MMLU Accuracy After Pruning.}
On \benchbg{acccolor}{HellaSwag} and \benchbg{acccolor}{MMLU}, compression induces smaller accuracy drops than on stereotype QA benchmarks, and Debias-SparseGPT remains comparable to SparseGPT (59.76\% vs.\ 59.11\% MMLU on LLaMA; 67.73\% vs.\ 67.35\% on Qwen). \benchbg{pplcolor}{Perplexity} on Wiki-2 follows the same pattern relative to dense models, with magnitude pruning affecting LLaMA and Qwen more strongly ($\sim$9.5).

\paragraph{Predictive Uncertainty Explains Variability in Improvements Across Models.}
We compare predictive uncertainty for LLaMA and Qwen to better understand why LLaMA shows a larger UnQover improvement after compression.
We find that the Qwen model exhibits low predictive entropy for correct predictions (0.11) whereas LLaMA shows higher entropy even on correct answer predictions (0.94).
We report these evaluation results in \autoref{tab:qwen_confidence} (see \autoref{sec:extended-results}).
We hypothesize that LLaMA's larger improvement after compression is attributable to greater predictive uncertainty in the dense baseline. This interpretation is consistent with \citet{proskurina-etal-2024-quantization}, who report that shifts in confidence distributions are driven primarily by instances that are uncertain under the dense model.

\begin{table*}[t]
\centering
\small
\setlength{\tabcolsep}{6pt}
\renewcommand{\arraystretch}{1.15}

\begin{tabular}{ll A O D}
\toprule
\textbf{Method} &
\textbf{Calibration Data} &
\textbf{MMLU $\uparrow$} &
\textbf{UnQover $\uparrow$} &
\textbf{DTO $\downarrow$} \\
\midrule

\multirow{3}{*}{SparseGPT}
& StereoSet
& 50.41
& 28.45
& 0.616 \\
& + UltraChat
& 53.84
& 42.46
& 0.522 \\
& \textbf{Diff.}
& \textbf{+3.43}
& \textbf{+14.01}
& \textbf{-0.094} \\
\midrule

\multirow{3}{*}{\textbf{Debias-SparseGPT}}
& StereoSet
& 48.16
& 24.94
& 0.645 \\
& \textbf{+ UltraChat}
& \textbf{54.17}
& \textbf{47.26} 
& \textbf{0.494} \\ 
& \textbf{Diff.}
& \textbf{+6.01}
& \textbf{+22.32} 
& \textbf{-0.151} \\  
\bottomrule
\end{tabular}

\caption{Effect of adding UltraChat calibration at \texttt{2:4} sparsity level. Diff. denotes the change relative to StereoSet-only calibration at the same sparsity level for the corresponding method, as reported in \autoref{tab:qwen-sparsity-all}.}
\label{tab:qwen_ultrachat_effect}
\end{table*}

\subsection{Debias-SparseGPT Across Sparsity Regimes}
Next, we assess whether the observed performance generalizes to higher sparsity levels and pruning regimes.
\autoref{tab:qwen-sparsity-all} reports evaluation results for Qwen-Instruct models compressed under unstructured sparsity levels of 25\% and 50\%, and semi-structured 2:4 restricted sparsity.
Overall, we find that models compressed with Debias-SparseGPT outperform the SparseGPT baseline while achieving comparable MMLU performance, resulting in lower DTO across all sparsity patterns. 
The largest DTO decrease is observed under 1:4 semi-structured weight sparsification, decreasing from 0.311 to 0.291 (\autoref{tab:debias-sparsegpt}).

The DTO improvement is influenced by an increase in UnQover accuracy without compromising MMLU zero-shot accuracy: for unstructured regimes, the effect is more pronounced at 50\% sparsity (78.35 to 80.35), while for semi-structured regimes it is most pronounced at 1:4 (70.60 to 74.41).
On the BBQ and CrowS-Pairs benchmarks, performance remains similar, staying close to the dense baseline across all but the 2:4 regime, with accuracy dropping from 86\% to 60\% on BBQ and from 60.8 to 57.7 on CrowS-Pairs.

\paragraph{Calibration data impact at \texttt{2:4} sparsity}
We use StereoSet as a calibration corpus; however, it is limited in diversity, comprising only \(\sim\)4k unique tokens. 
Consequently, under more aggressive pruning regimes such as \texttt{2:4}, perplexity nearly doubles (\autoref{tab:qwen-sparsity-all}), and the UnQover score approaches the random baseline of 33\%.
To mitigate this effect, we perform additional experiments with UltraChat~\citep{ding-etal-2023-enhancing}, which contains dialogues covering a wide range of topics.
We augment StereoSet with 256 UltraChat examples (\(\sim\)15k tokens) to improve calibration coverage.
For Debias-SparseGPT, the Hessian in Eq.~\eqref{eq:debias-hessian} is computed over the StereoSet pairs; for UltraChat, the Hessian is accumulated without the \(\Delta \mX\) term, since UltraChat texts are unpaired.
This setting allows for evaluating the isolated benefit of the \(\Delta \mX\) term in the Hessian.
We report the evaluation results for these experiments in \autoref{tab:qwen_ultrachat_effect}.

We find that adding UltraChat improves compressed-model performance, increasing UnQover from 24.9 to 47.3 and MMLU from 48.16 to 54.17, with a corresponding decrease in DTO.
BBQ accuracy also increases for Debias-SparseGPT, from 66.70 to 78.60, and Debias-SparseGPT outperforms SparseGPT on all three fairness benchmarks (see \autoref{tab:debias-sparsegpt-ultrachat-appendix}).
Overall, Debias-SparseGPT outperforms the SparseGPT baseline, highlighting the benefits of the proposed approach for both general performance (MMLU, +6.01) and bias reduction (UnQover, +22.32), with larger gains in the latter.
In \autoref{app:calibration-data-choice-rebuttal}, we further examine how StereoSet category-specific calibration affects these results.

\subsection{Efficiency Evaluation}
Finally, we compare the efficiency of models compressed with SparseGPT and Debias-SparseGPT.
For a weight matrix $\mW\in\mathbb{R}^{n\times d}$, pruning memory is dominated by the Hessian, the Cholesky factor, the weights, and the residual matrix. Since the bias-aware term is accumulated directly into $\mH$, Debias-SparseGPT preserves SparseGPT's layer-wise memory complexity, $\mathcal{O}(d^2+nd)$, or $\mathcal{O}(d^2)$ for the transformer projections considered in our experiments.
\begin{table}[h]
\centering
\scriptsize{
\begin{tabular}{l D A A}
\toprule
\textbf{Model} 
& \textbf{DTO} $\downarrow$
& \textbf{Throughput} $\uparrow$
& \textbf{CO$_2$} $\downarrow$ \\
&  
& \textbf{(tok/s)} 
& \textbf{(kg/Mtok)} \\
\midrule
Qwen-2.5-7B (Dense)  
& 0.291  & 27.54 & 0.0998  \\

SparseGPT (\texttt{2:4}) 
& 0.522 & \textbf{73.05} & \textbf{0.0376} \\

\textbf{Debias-SparseGPT (\texttt{2:4})} 
& \textbf{0.494} & \textbf{73.05} & \textbf{0.0376} \\
\bottomrule
\end{tabular}}
\caption{
Efficiency evaluation results for Qwen-2.5-7B under 2:4 semi-structured sparsity.
}\label{table:mfu_sparsity_results}
\end{table}

We evaluate Qwen-2.5-7B and its pruned variants under 2{:}4 semi-structured sparsity using vLLM on an NVIDIA A100 GPU with Tensor Core sparsity support~\citep{kwon2023efficient}.
Results are reported in \autoref{table:mfu_sparsity_results}.
Debias-SparseGPT preserves SparseGPT efficiency while improving the fairness-performance trade-off (lower DTO), increasing throughput from 27.54 to 73.05 tokens/s.

\section{Conclusion}\label{sec:conclusion}
In this paper, we present \textbf{Debias-SparseGPT}, an approach to model compression aimed at reducing the disparate impact of pruning on model performance in causal question answering. 

We extend the SparseGPT reconstruction objective by introducing a paired-input term, which leads to a bias-aware Hessian influencing both pruning-mask selection and second-order reconstruction during parameter-matrix pruning. 
We apply this approach to nine LLMs from diverse model families, including instruction-tuned models, and observe consistent improvements on \textsc{UnQover} and \textsc{BBQ} under semi-structured and unstructured sparsity levels. 
Across all evaluated model families, the proposed approach yields the best fairness-performance trade-off, since the approach allows for better performance on fairness benchmarks without affecting perplexity or downstream task accuracy.

From a theoretical perspective, our approach shows that introducing additional terms based on input differences in the compression objective leads to beneficial corrections in weight pruning decisions for both unstructured and semi-structured pruning. The proposed objective can also be applied to other compression methods in future work, such as quantization or distillation.

\section*{Limitations}

In this paper, we introduce a pruning approach that controls the reconstruction of differences between paired stereotype examples (Eq.~\eqref{eq:fair-sparse-objective}) and validate it across models and sparsity levels; nevertheless, several limitations should be noted.

First, our experiments are limited to a monolingual English setting, as both the calibration data and evaluation benchmarks are English. Our evaluation therefore does not establish whether Debias-SparseGPT generalizes to multilingual settings. In addition, recent studies suggest that using multilingual calibration data can further improve the performance of compressed models \citep{williams-aletras-2024-impact}.   

Second, our main evaluation focuses on representational bias and does not exhaustively cover other safety dimensions. 
Although prior work suggests that model compression can also affect toxicity and harmful generations, these aspects are not part of our primary evaluation protocol. 
We therefore conduct an additional safety analysis on RealToxicityPrompts and HarmBench, reported in \autoref{app:safety-evaluation}. 
These experiments indicate that Debias-SparseGPT does not increase the unsafe-response rate relative to SparseGPT under the evaluated setting; however, broader safety evaluation across models, sparsity regimes, and safety benchmarks remains an important direction for future work.

Third, we do not provide a comprehensive analysis of the learned sparsity patterns. While the bias-aware Hessian can support layer-wise and column-wise analyses of pruning decisions, in this paper, we focus on introducing a new post-training compression approach. We provide an initial analysis of the resulting sparsity patterns in \autoref{app:sparsity-pattern-analysis} and leave a more detailed investigation for future work.

Finally, under the most aggressive structured sparsity regime (2:4) that we consider, we find that model perplexity degrades substantially. 
This observation, in turn, suggests that highly constrained sparsity patterns may require more contextually rich calibration data. 
Our experiments with UltraChat demonstrate that longer and more diverse texts can improve performance (see \autoref{app:calibration-data-choice-rebuttal}).
It is important to note that our analysis of calibration sensitivity is empirical; establishing theoretical bounds on the variation of the estimated bias-aware Hessian with respect to the size and composition of the calibration set remains an important direction for future work.
Further improvements of compressed models can be obtained through adapted sparse training \citep{huang2025pruning}, parameter-efficient fine-tuning, or regional optimization of the retained weights \citep{yang-etal-2025-wanda}.
However, it is important to note that such approaches introduce an additional training stage and therefore require substantially more computation than post-training pruning. 
Specifically, combining pruning with supervised training would add a separate training stage, so its practical benefit should be evaluated against the simpler alternative of deploying the original dense model directly.
In contrast, our goal is to study whether bias amplification can be mitigated directly during compression, without increasing the computational cost relative to SparseGPT. 

\section*{Ethical Considerations}

\paragraph{Usage of Scientific Artifacts}
In this paper, we experiment with six datasets, and our usage complies with the intended research purposes of these benchmarks. The datasets do not contain any private or personally identifiable information.
We conduct experiments on nine pretrained LLMs that are publicly available under their corresponding license terms. Several models are distributed under standard permissive licenses.
Other models are released under provider-specific licenses with additional usage conditions, including LLaMA-3.1-8B-IT,\footnote{\url{https://www.llama.com/llama3_1/license/}} Vicuna-7B-v1.5-IT,\footnote{\url{https://ai.meta.com/llama/license/}} and Gemma-2-9B-IT.\footnote{\url{https://ai.google.dev/gemma/terms}} 
Our usage of all models complies with their respective license terms, including the requirements governing the distribution of model derivatives, under which compressed models fall, as specified in the corresponding agreements.
We list all license information in \autoref{app:experimental-settings}.

\paragraph{Intended Use}
We integrate Debias-SparseGPT into the \texttt{llm-compressor} package, which supports a broad range of large language and multimodal models and will be released under the Apache-2.0 license upon acceptance. 
Our implementation \footnote{\href{https://github.com/upunaprosk/debias-llm-compressor}{github.com/upunaprosk/debias-llm-compressor}} is compatible with other modifiers in the framework, allowing the Debias-SparseGPT sparsification step to be combined with quantization and other compression techniques for additional efficiency gains.

\paragraph{Potential Risks}
Releasing compressed models and code for reproducibility may enable misuse, including the generation of stereotyped, biased, or toxic content targeting specific communities. 
Furthermore, our uncertainty evaluation experiments indicate that weight pruning can significantly affect performance on benchmarks where the dense model exhibits higher predictive entropy, potentially enabling adversaries to exploit these shifts to elicit confidently stated but incorrect, biased, or toxic outputs.

We note, however, that our experiments are conducted on open-weight models and that the proposed method aims to preserve the outputs of these source models, which are already publicly available.
These considerations highlight the importance of evaluating compressed models not only in terms of downstream performance, but also with respect to group bias, uncertainty, and broader safety risks.

\section*{Acknowledgments}

This work was supported by the French National Research Agency through the \textbf{ANR-Diké project}\footnote{\url{https://www.anr-dike.fr}} (ANR-21-CE23-0026).
The experiments presented in this work were conducted using HPC resources provided by GENCI-IDRIS (Grant 2025-AD011014384R1).

\bibliography{custom}

\onecolumn
\appendix

\section{Notations}\label{sec:notations}
We provide a list of notations used throughout the paper in \autoref{tab:notation}, presented separately for the metric definitions and the Debias-SparseGPT method.

\begin{table*}[h]
\centering
\begin{tabular}{p{0.24\textwidth} p{0.70\textwidth}}
\toprule
\textbf{Symbol} & \textbf{Definition} \\
\midrule

$\mathcal{M}$ & Language model under evaluation or compression \\
$x$ & Input instance (e.g., question or prompt) \\
$\mathcal{D}$ & Evaluation dataset \\
$\mathcal{G}$ & Group category (e.g., gender, race, religion, nationality, \dots) \\
$\mathcal{D}_{\mathcal{G}}$ & Subset of $\mathcal{D}$ associated with group category $\mathcal{G}$ \\
$\mathcal{A}_{\text{unk}}(\mathcal{M},\mathcal{G})$ & Unknown-answer accuracy for group $\mathcal{G}$ \\
\midrule

$\mW \in \mathbb{R}^{n\times d}$ & Dense weight matrix before pruning \\
$\mW' \in \mathbb{R}^{n\times d}$ & Weight matrix after second-order compensation and before masking \\
$\mM \in \{0,1\}^{n\times d}$ & Binary sparsity mask, where $1$ denotes a retained weight and $0$ a pruned weight \\
$\tilde{\mW}=\mW'\odot\mM \in \mathbb{R}^{n\times d}$ & Masked weight matrix \\
$\hat{\mW} \in \mathbb{R}^{n\times d}$ & Pruned weight matrix \\
$\Delta\mW=\hat{\mW}-\mW \in \mathbb{R}^{n\times d}$ & Change in the weight matrix induced by pruning and reconstruction \\

$\mX_0,\mX_1 \in \mathbb{R}^{d\times N_{t}}$ & Pro-/anti-stereotypical paired input representations of $N_{t}$ tokens \\
$\Delta\mX=\mX_0-\mX_1 \in \mathbb{R}^{d\times  N_{t}}$ & Difference between paired input representations \\
$\vw=\operatorname{vec}_r(\mW) \in \mathbb{R}^{nd}$ & Row-wise vectorization of $\mW$ \\
$\hat{\vw}=\operatorname{vec}_r(\hat{\mW}) \in \mathbb{R}^{nd}$ & Row-wise vectorization of the pruned weight matrix \\
$\Delta\vw=\hat{\vw}-\vw \in \mathbb{R}^{nd}$ & Vectorized weight update \\
$\mJ_{\vw} \in \mathbb{R}^{nd}$ & Gradient of the reconstruction objective with respect to $\vw$ \\
$\mH_{\vw} \in \mathbb{R}^{nd\times nd}$ & Weight-space Hessian of the reconstruction objective \\
$\mH \in \mathbb{R}^{d\times d}$ & Input-space Hessian, $\mX_0\mX_0^\top+\mX_1\mX_1^\top+2\Delta\mX\Delta\mX^\top$ \\
$\mI_n \in \mathbb{R}^{n\times n}$ & Identity matrix \\
$\ve_p \in \mathbb{R}^{nd}$ & $p$-th canonical basis vector \\
$w_p$ & $p$-th entry of $\vw$ \\
$\varepsilon_p$ & Saliency of weight $w_p$, $\varepsilon_p=\dfrac{w_p^2}{2[\mH_{\vw}^{-1}]_{pp}}$ \\

$s$ & Target sparsity ratio, i.e., the fraction of weights pruned \\
$B$ & Number of columns processed in each pruning block \\
$B_s$ & Saliency stride, with $B_s=M$ under $N\!:\!M$ semi-structured sparsity \\

$\mH_{\text{acc}} \in \mathbb{R}^{d\times d}$ & Reconstruction Hessian term, $\mX_0\mX_0^\top+\mX_1\mX_1^\top$ \\
$\mH_{\text{bias}} \in \mathbb{R}^{d\times d}$ & Bias-aware Hessian term, $2(\mX_0-\mX_1)(\mX_0-\mX_1)^\top$ \\
$\mC \in \mathbb{R}^{d\times d}$ & Cholesky factor satisfying $\mC^\top\mC=\mH^{-1}$ \\
$\mE \in \mathbb{R}^{n\times B}$ & Block-wise residual matrix used to propagate second-order compensation \\
$\lambda$ & Hessian damping coefficient \\

\bottomrule
\end{tabular}

\caption{
Summary of notation used for the evaluation metrics and the Debias-SparseGPT formulation in \S\ref{sec:debias-sparsegpt-method}.
}
\label{tab:notation}
\end{table*}

\section{Detailed Theoretical Solution}
\label{app:detailed-solution}

In this appendix, we provide a detailed derivation of the Debias-SparseGPT weight update induced by the proposed debiasing objective in Eq.~\eqref{eq:fair-sparse-objective}.

We denote the weight update and the paired-input difference by
$\Delta\mW=\hat{\mW}-\mW$ and $\Delta\mX=\mX_0-\mX_1$, respectively.
Let $\vw=\operatorname{vec}_r(\mW)$,
$\hat{\vw}=\operatorname{vec}_r(\hat{\mW})$, and
$\Delta\vw=\hat{\vw}-\vw=\operatorname{vec}_r(\Delta\mW)$.
Using a second-order Taylor expansion of the layer-wise objective $f$
around the pretrained parameters $\vw$ \citep{hassibi1993optimal}, we obtain:
\begin{equation}
\label{eq:sparse-taylor}
f(\vw+\Delta\vw)\approx f(\vw)+\mJ_{\vw}^\top\Delta\vw+\frac{1}{2}\Delta\vw^\top\mH_{\vw}\Delta\vw,
\end{equation}
where $\mJ_{\vw}$ and $\mH_{\vw}$ denote the gradient and Hessian with respect to $\vw$.

At the pretrained weights, we assume first-order stationarity, i.e., $\mJ_{\vw}=\mathbf{0}$, since the model parameters have already been optimized during training. Under this assumption, only the quadratic term remains. For the debiasing objective in Eq.~\eqref{eq:fair-sparse-objective}, the Hessian takes the form $\mH_{\vw}=\mI_n\otimes\mH$, where $\mI_n$ is the $n\times n$ identity matrix, and the input-space Hessian is:
\begin{equation}
\label{app-eq:debias-hessian}
\mH=\mX_0\mX_0^\top+\mX_1\mX_1^\top+2\,\Delta\mX\,\Delta\mX^\top.
\end{equation}

Pruning a single weight at index $p$ enforces $(\hat{w})_p=0$, or equivalently $\ve_p^\top\Delta\vw=-w_p$, where $\ve_p$ denotes the $p$-th canonical basis vector and $w_p$ is the $p$-th entry of $\vw$. This yields the constrained problem:
\begin{equation}
\label{eq:sparse-opt-prob}
\min_{\Delta\vw}\;\frac{1}{2}\Delta\vw^\top\mH_{\vw}\Delta\vw
\quad \text{s.t.} \quad
\ve_p^\top\Delta\vw=-w_p.
\end{equation}

To solve Eq.~\eqref{eq:sparse-opt-prob}, we introduce the Lagrangian:
\[
L(\Delta\vw,\mu)=\frac{1}{2}\Delta\vw^\top\mH_{\vw}\Delta\vw+\mu(\ve_p^\top\Delta\vw+w_p).
\]
Setting the derivative with respect to $\Delta\vw$ to zero gives $\mH_{\vw}\Delta\vw+\mu\ve_p=0$. Solving for the update and enforcing the constraint yields the Lagrange multiplier $\mu=w_p/[\mH_{\vw}^{-1}]_{pp}$. Substituting this back gives the optimal perturbation:
\begin{equation}
\label{eq:app-sparse-solution-proof}
\Delta\vw^{\star}=-\frac{w_p}{[\mH_{\vw}^{-1}]_{pp}}\,\mH_{\vw}^{-1}\ve_p.
\end{equation}

The corresponding increase in the objective, which serves as the saliency $\varepsilon_p$ (as in the OBS framework; \citealp{hassibi1993optimal}), is:
\begin{equation}
\label{app-eq:sparse-saliency-proof}
\varepsilon_p=\frac{1}{2}(\Delta\vw^\star)^\top\mH_{\vw}\Delta\vw^\star
=\frac{w_p^2}{2[\mH_{\vw}^{-1}]_{pp}}.
\end{equation}
Larger values of $\varepsilon_p$ correspond to a larger increase in the objective function upon removal of coordinate $p$. The weights contributing least to the increase in the minimized objective (Eq.~\eqref{eq:fair-sparse-objective}) are selected for pruning in the pruning mask $\mM$.

Overall, the theoretical solution for the change in the weight matrix $\mW$, defined in Eq.~\eqref{eq:app-sparse-solution-proof}, for the Debias-SparseGPT objective in Eq.~\eqref{eq:fair-sparse-objective}, depends on the weight-space Hessian computed over input representation pairs of pro-stereotypical and anti-stereotypical texts. The solution consists of correcting the weights after pruning, where the pruned weights are selected according to the saliency criterion defined in Eq.~\eqref{app-eq:sparse-saliency-proof}.

The resulting optimization is summarized in \autoref{fig:debias_sparsegpt_mask_construction} and Algorithm~\ref{alg:debias-sparsegpt}. In particular, the bias-aware Hessian in Eq.~\eqref{app-eq:debias-hessian} is used both to define the saliency criterion for pruning-mask construction and to compute the second-order correction of the retained weights. Consequently, the additional term defined over paired input differences affects both the selection of pruned parameters and the subsequent reconstruction of the weight matrix.

\clearpage
\twocolumn
\begin{table*}[h]
\centering
\resizebox{0.98\textwidth}{!}{%
\begin{tabular}{lcccc}
\toprule
\textbf{Model} & \textbf{Params} & \textbf{IT} & \textbf{Multilingual} & \textbf{Link} \\
\midrule
LLaMA-3.1-8B-IT & 8B   & $\checkmark$ & $\checkmark$ & \href{https://hf.co/meta-llama/Llama-3.1-8B-Instruct}{hf.co/meta-llama/Llama-3.1-8B-Instruct} \\
Qwen-2.5-7B-IT & 7B   & $\checkmark$ & $\checkmark$ & \href{https://hf.co/Qwen/Qwen2.5-7B-Instruct}{hf.co/Qwen/Qwen2.5-7B-Instruct} \\
Vicuna-7B-v1.5-IT & 7B   & $\checkmark$ & $\times$     & \href{https://hf.co/lmsys/vicuna-7b-v1.5}{hf.co/lmsys/vicuna-7b-v1.5} \\
Aya-Expanse-8B & 8B   & $\checkmark$ & $\checkmark$ & \href{https://hf.co/CohereLabs/aya-expanse-8b}{hf.co/CohereLabs/aya-expanse-8b} \\
Gemma-2-9B-IT & 9B   & $\checkmark$ & $\times$     & \href{https://hf.co/google/gemma-2-9b-it}{hf.co/google/gemma-2-9b-it} \\
Mistral-7B-v0.3-IT & 7B   & $\checkmark$ & $\times$     & \href{https://hf.co/mistralai/Mistral-7B-Instruct-v0.3}{hf.co/mistralai/Mistral-7B-Instruct-v0.3} \\
Phi-4-Mini-IT & 3.8B & $\checkmark$ & $\checkmark$ & \href{https://hf.co/microsoft/Phi-4-mini-instruct}{hf.co/microsoft/Phi-4-mini-instruct} \\
\midrule
Deepseek-LLM-7B (Base) & 7B   & $\times$     & $\times$     & \href{https://hf.co/deepseek-ai/deepseek-llm-7b-base}{hf.co/deepseek-ai/deepseek-llm-7b-base} \\
Qwen-3-8B (Base) & 8B   & $\times$     & $\checkmark$ & \href{https://hf.co/Qwen/Qwen3-8B}{hf.co/Qwen/Qwen3-8B} \\
\bottomrule
\end{tabular}}
\caption{Instruction-tuned and base models used in our experiments, together with their parameter counts and multilingual support. IT denotes instruction-tuned models.}
\label{tab:models-overview-app}
\end{table*}

\section{Experimental Settings}\label{app:experimental-settings}

In this appendix, we provide additional details on the experimental setup and implementation.

\paragraph{Models}
We run our experiments on nine LLMs, including both base and instruction-tuned models, which are listed with links in \autoref{tab:models-overview-app}.
Our usage of models complies with their respective licenses, which are listed in \autoref{tab:model-licenses}. 
License files are available at the official model repository links.
The compression produces derivative versions of the original models through weight pruning. 
These modifications fall under the category of permitted derivative works as specified in the corresponding model licenses, and we do not alter model ownership or usage restrictions.

\begin{table*}[h]
\centering
\begin{tabular}{ll}
\toprule
\textbf{Model} & \textbf{License} \\
\midrule
LLaMA-3.1-8B-IT & LLaMA 3.1 Community License \\
Qwen-2.5-7B-IT & Apache-2.0 \\
Vicuna-7B-v1.5-IT & LLaMA 2 Community License \\
Aya-Expanse-8B & CC BY-NC \\
Gemma-2-9B-IT & Gemma Terms of Use (Google) \\
Mistral-7B-v0.3-IT & Apache-2.0 \\
Phi-4-Mini-IT & MIT \\
\midrule
Deepseek-LLM-7B (Base) & DeepSeek License \\
Qwen-3-8B (Base) & Apache-2.0 \\
\bottomrule
\end{tabular}
\caption{License types for all pretrained models used in our experiments. Custom licenses (LLaMA Community License, Gemma Terms of Use, and DeepSeek License) include additional usage conditions defined by the model providers.}
\label{tab:model-licenses}
\end{table*}

\paragraph{Stereotype Evaluation Benchmarks}
We conduct experiments on three benchmarks designed to evaluate stereotypes in LLMs, covering different target groups.

The \textbf{BBQ} benchmark \citep{parrish-etal-2022-bbq} consists of question-answering tasks that prompt responses referring to target groups, often under insufficient contextual information. 
We use the group-balanced \textbf{BBQ} set released for the HELM leaderboard \citep{helm-safety} with the original BBQ metric implementation.

The \textbf{UnQover} benchmark \citep{li-etal-2020-unqovering} follows a structure similar to BBQ and includes contextualized questions spanning four attributes. We use the implementation provided by \citet{sun-etal-2024-causal}.
For both \textbf{UnQover} and \textbf{BBQ}, the evaluation metric is defined as the accuracy of predictions selecting the \emph{Unknown} or \emph{Not determined} answer option.
Following the source implementation, predictions are obtained by scoring the candidate answer tokens using the next-token logits. 

The \textbf{CrowS-Pairs} benchmark \citep{nangia-etal-2020-crows} consists of minimal pairs of stereotypical and anti-stereotypical sentences. 
The metric is defined as the percentage of cases in which the stereotypical sentence is assigned a higher likelihood than the anti-stereotypical one: $\text{Likelihood}(x_{\text{stereo}}) > \text{Likelihood}(x_{\text{anti}})$.
The ideal CrowS-Pairs score is $0.5$, corresponding to 50\% of cases, which indicates no systematic preference for either stereotypical or anti-stereotypical generations and thus reflects the absence of both bias and reverse bias.

An overview of the benchmarks used is provided in \autoref{tab:fairness_benchmarks}.
\begin{table*}[!t]
    \centering
    
    \resizebox{0.98\textwidth}{!}{%
    \begin{tabular}{p{0.3\linewidth} c p{0.6\linewidth}}
    \toprule
    \textbf{Benchmark} & \textbf{Size} & \textbf{Attributes (target groups)} \\
    \midrule
    BBQ-HELM~\cite{parrish-etal-2022-bbq} 
    & 27,000 
    & Age, disability status, gender identity, nationality, physical appearance, ethnicity, religion, socio-economic status, sexual orientation \\

    CrowS-Pairs~\cite{nangia-etal-2020-crows} 
    & 3,016  
    & Race, gender, socio-economic status/occupation, nationality, religion, age, sexual orientation, physical appearance, disability \\

    UnQover~\cite{li-etal-2020-unqovering} 
    & 40,000 
    & Gender, nationality, ethnicity, religion \\
    \bottomrule
    \end{tabular}}
    \caption{Overview of benchmarks used to evaluate stereotyping bias in language models.}
    \label{tab:fairness_benchmarks}
\end{table*}

\paragraph{Calibration Data and Pruning Configuration}
For the main pruning experiments, we use the StereoSet development set as calibration data, comprising 4,212 paired pro- and anti-stereotypical examples.
The same calibration data are used for Wanda, SparseGPT, and Debias-SparseGPT to ensure a consistent comparison across data-dependent pruning methods.
Under the 2:4 sparsity setting, we additionally augment StereoSet with 256 UltraChat examples to increase the contextual diversity of the calibration set.
For Debias-SparseGPT, the bias-aware term is computed only over the paired StereoSet examples, while UltraChat examples \citep{ding-etal-2023-enhancing} contribute to the reconstruction Hessian without the paired-input difference term.
We evaluate unstructured sparsity levels of 25\% and 50\%, as well as 1:4 and 2:4 semi-structured sparsity.

\paragraph{Debias-SparseGPT Implementation}
All experiments are conducted on two NVIDIA A100 GPUs with 80 GB of memory each.
Implementations of the Magnitude, Wanda, and SparseGPT baselines are based on the \texttt{LLM-Compressor} package \citep{llmcompressor2024}.
Debias-SparseGPT is implemented within the same framework and preserves the layer-wise memory complexity of SparseGPT, since the bias-aware term is accumulated directly into the Hessian without requiring an additional $d\times d$ matrix.
We use the following default settings from the framework: a block size of 128 and a Hessian damping fraction of 0.01, unless otherwise specified.
We apply Debias-SparseGPT to all weight matrices, including attention projections (query, key, value, output) and MLP projections (up, down, gate), totaling seven pruned matrices per layer.
For each pro-/anti-stereotypical calibration pair, we verify that the tokenized sequences have equal length and differ only in the demographic-group substitution, ensuring positional alignment when computing $\Delta\mX=\mX_0-\mX_1$.
For all data-dependent pruning methods, we use the same calibration data to ensure a consistent comparison.
We make the implementation of Debias-SparseGPT openly available.\footnote{\href{https://github.com/upunaprosk/debias-llm-compressor}{github.com/upunaprosk/debias-llm-compressor}}

\paragraph{Evaluation Setup}
For the multiple-choice general-performance benchmarks HellaSwag and MMLU, we use the LM Evaluation Harness implementation.\footnote{\href{https://github.com/EleutherAI/lm-evaluation-harness}{github.com/EleutherAI/lm-evaluation-harness}}
The model is evaluated by scoring the logits of the next token for each candidate answer.

For the efficiency evaluation, we additionally report the estimated carbon footprint per 1M generated tokens.
Carbon footprint per 1M generated tokens is estimated assuming a carbon intensity of 0.033\,kgCO$_2$e/kWh, using the Optimum-Benchmark implementation.\footnote{\href{https://github.com/huggingface/optimum-benchmark}{github.com/huggingface/optimum-benchmark}}

\section{Extended Results}
\label{sec:extended-results}

In this appendix, we provide extended evaluation results complementing the experiments in \S\ref{sec:results-comparison-to-baselines}. We first extend the comparison under 1:4 semi-structured sparsity to six additional model families and then provide category-level and uncertainty evaluation for selected models. 
We additionally report results across the complete set of sparsity regimes and provide a detailed comparison under the more restrictive 2:4 setting.

\paragraph{Results Across Additional Model Families}
\autoref{tab:debias-sparsegpt-newmodels} reports performance and stereotype generation evaluation results for six models, compared against the SparseGPT and Wanda baselines using the same calibration data under a 1:4 sparsity setting. 
Across all six models, models compressed with Debias-SparseGPT achieve higher UnQover accuracy and lower DTO scores relative to SparseGPT while maintaining comparable perplexity and downstream-task performance.
The largest UnQover improvement among these models is observed for Phi-4-Mini, increasing from 43.36\% to 48.51\%, followed by Qwen3-8B, where accuracy increases from 61.84\% to 65.00\%.
For DeepSeek-8B, where the dense model already exhibits low UnQover accuracy, compression with Debias-SparseGPT nevertheless improves accuracy relative to SparseGPT, from 17.66\% to 20.08\%.

\begin{figure}[t]
    \centering
    \includegraphics[width=0.98\linewidth]{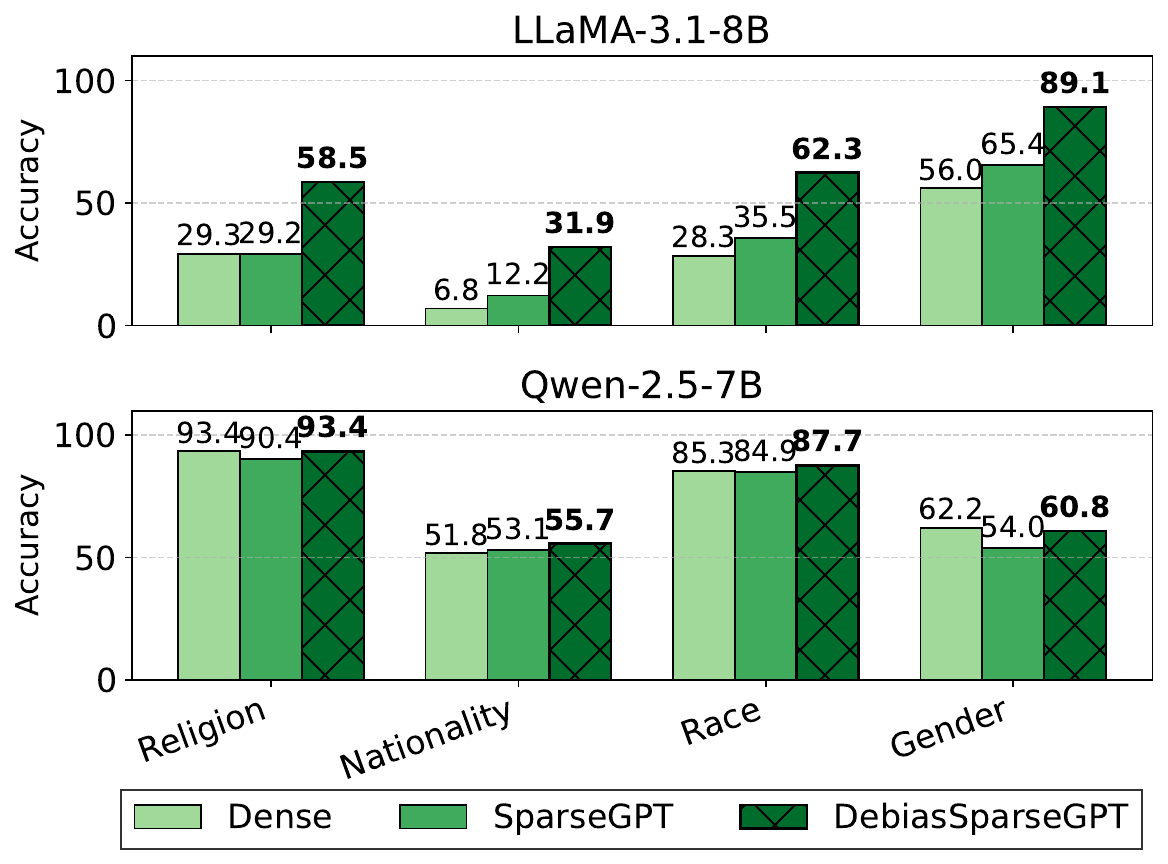}
    \caption{Per-category accuracy (\%) across models on the \benchbg{othercolor}{\strut UnQover} benchmark. 
    Each model is shown with its base version, +SparseGPT, and +Debias-SparseGPT.}
    \label{fig:accuracy-heatmap}
\end{figure}

\paragraph{Category-Level UnQover Evaluation}
\autoref{fig:accuracy-heatmap} shows category-wise accuracy on the UnQover benchmark for the Qwen-2.5 and LLaMA-3.1-8B models compressed with SparseGPT and Debias-SparseGPT. The models are evaluated separately on the \benchbg{othercolor}{UnQover} benchmark for each question category (Religion, Nationality, Race, and Gender). 
For the LLaMA-3.1-8B model, compression with Debias-SparseGPT results in higher UnQover accuracy than SparseGPT across all four evaluated categories. The largest improvements are observed for religion, increasing from 29.2\% to 58.5\%, and race, increasing from 35.5\% to 62.3\%. 
For the Qwen-2.5-7B model, compression with Debias-SparseGPT results in higher accuracy across all four categories as well, with the largest improvements observed for gender (from 54.0\% to 60.8\%) and religion (from 90.4\% to 93.4\%).

\begin{table}[h]
\centering
\footnotesize
\begin{tabular}{lcc}
\toprule
\textbf{Metric} & \textbf{Qwen-2.5-7B} & \textbf{LLaMA-3.1-8B} \\
\midrule
Accuracy & 73.17 & 30.10 \\
\midrule
Confidence (Correct) & 0.960 & 0.545 \\
Confidence (Wrong)   & 0.874 & 0.600 \\
\midrule
Entropy (Correct)    & 0.109 & 0.938 \\
Entropy (Wrong)      & 0.308 & 0.858 \\
\bottomrule
\end{tabular}
\caption{Prediction confidence and predictive entropy statistics on the \benchbg{othercolor}{\strut UnQover} benchmark. Qwen exhibits higher confidence for correct predictions and lower entropy, indicating lower predictive uncertainty than LLaMA.}
\label{tab:qwen_confidence}
\end{table}

\paragraph{Predictive Uncertainty}
\autoref{tab:qwen_confidence} reports predictive confidence and entropy on UnQover for the dense Qwen-2.5-7B and LLaMA-3.1-8B models.
Predictive entropy is computed from the model probabilities over candidate answers as
$H(p)=-\sum_i p_i\log p_i$.
We find that Qwen exhibits substantially higher confidence and lower entropy for correct predictions than LLaMA, with confidence values of 0.960 and 0.545, respectively.
A similar difference is observed for incorrect predictions, with entropy of 0.308 for Qwen and 0.858 for LLaMA.
Overall, these evaluation results indicate that the dense LLaMA model has lower confidence in the answer selection on the UnQover benchmark.
This observation is consistent with the larger changes observed in LLaMA after pruning and with our hypothesis that examples for which the dense model is less confident are more susceptible to compression-induced changes.

\paragraph{Results Across Sparsity Regimes}
\autoref{tab:debias-sparsegpt-all-models-appendix} extends the sparsity analysis to LLaMA, Vicuna, and Qwen under 25\% and 50\% unstructured sparsity and 1:4 and 2:4 semi-structured sparsity.
We find that across less-restrictive sparsity settings, Debias-SparseGPT generally preserves SparseGPT downstream performance while improving UnQover accuracy.
The largest improvement is observed for LLaMA under 1:4 sparsity, where UnQover increases from 35.60\% to 60.46\%.
Under 2:4 sparsity, performance degrades more strongly for all models, further motivating the calibration-data analysis presented in \autoref{app:calibration-data-choice-rebuttal}.

\begin{table*}[t!]
\centering
\resizebox{0.98\textwidth}{!}{%
\begin{tabular}{l P A A O O O D}
\toprule
\textbf{Model + Method} 
& \textbf{PPL $\downarrow$} 
& \textbf{HellaSwag $\uparrow$} 
& \textbf{MMLU $\uparrow$}
& \textbf{UnQover $\uparrow$} 
& \textbf{BBQ $\uparrow$} 
& \textbf{CP $\downarrow$}
& \textbf{DTO $\downarrow$} \\
\midrule
\multicolumn{8}{c}{\textbf{Instruction-tuned Models}} \\
\midrule
\quad Aya-Expanse-8B
& 8.86 & 60.26 & 62.14 & 33.34 & 55.9 & 66.79 & 0.542 \\
\cdashlinelr{1-8}
\quad + Wanda           
& 10.18 & 57.14 & 57.88 & 20.97 & 51.8 & 65.65 & 0.633 \\
\quad + SparseGPT 
& 9.75 & 58.72 & 59.42 & 23.04 & 57.5 & 60.52 & 0.615 \\
\quad + \textbf{Debias-SparseGPT (ours)} 
& 9.79 & 58.96 & 59.69 & 23.66 & 55.1 & 60.47 & 0.610 \\
\midrule
\quad Mistral-7B-v0.3-IT
& 7.20 & 62.76 & 60.00 & 35.60 & 71.7 & 60.76 & 0.536 \\
\cdashlinelr{1-8}
\quad + Wanda           
& 7.62 & 60.18 & 57.81 & 27.51 & 62.9 & 61.12 & 0.593 \\
\quad + SparseGPT 
& 7.74 & 59.52 & 56.59 & 28.26 & 65.1 & 61.12 & 0.593 \\
\quad + \textbf{Debias-SparseGPT (ours)}
& 7.69 & 59.72 & 57.00 & \textbf{29.19}$^{\dagger}$ & 64.0 & 61.12 & 0.586 \\
\midrule
\quad Phi-4-Mini-IT
& 11.14 & 52.29 & 66.22 & 39.85 & 82.8 & 58.68 & 0.488 \\
\cdashlinelr{1-8}
\quad + Wanda           
& 13.72 & 49.37 & 59.49 & 45.06 & 66.5 & 58.44 & 0.483 \\
\quad + SparseGPT 
& 12.89 & 49.47 & 60.24 & 43.36 & 64.2 & 57.78 & 0.489 \\
\quad + \textbf{Debias-SparseGPT (ours)} 
& 12.98 & 49.87 & 60.33 & \textbf{48.51}$^{\dagger}$ & 67.3 & 58.38 & 0.460 \\
\midrule
\quad Gemma-2-9B-IT
& 10.09 & 53.47 & 33.19 & 74.03 & 90.8 & 61.30 & 0.507 \\
\cdashlinelr{1-8}
\quad + Wanda           
& 10.60 & 52.72 & 29.97 & 68.99 & 82.8 & 62.61 & 0.542 \\
\quad + SparseGPT 
& 10.13 & 52.63 & 30.24 & 79.04 & 89.3 & 61.30 & 0.515 \\
\quad + \textbf{Debias-SparseGPT (ours)} 
& 10.03 & 52.64 & 31.75 & 79.22 & 89.1 & 61.42 & 0.504 \\
\midrule
\multicolumn{8}{c}{\textbf{Base Models}} \\
\midrule
\quad Qwen-3-8B
& 10.79 & 57.10 & 72.90 & 66.74 & 90.3 & 60.23 & 0.303 \\
\cdashlinelr{1-8}
\quad + Wanda           
& 11.96 & 53.31 & 69.72 & 60.36 & 82.3 & 58.38 & 0.353 \\
\quad + SparseGPT 
& 11.71 & 53.87 & 69.65 & 61.84 & 85.4 & 60.83 & 0.345 \\
\quad + \textbf{Debias-SparseGPT (ours)} 
& 11.86 & 53.92 & 69.52 & \textbf{65.00}$^{\dagger}$ & 83.8 & 60.64 & 0.328 \\
\midrule
\quad Deepseek-LLM-7B
& 8.09 & 56.94 & 44.12 & 27.88 & 37.7 & 66.67 & 0.645 \\
\cdashlinelr{1-8}
\quad + Wanda           
& 9.08 & 54.31 & 39.69 & 21.94 & 36.6 & 65.41 & 0.698 \\
\quad + SparseGPT 
& 8.72 & 54.54 & 40.64 & 17.66 & 33.0 & 68.07 & 0.718 \\
\quad + \textbf{Debias-SparseGPT (ours)} 
& 8.67 & 54.50 & 42.14 & \textbf{20.08}$^{\dagger}$ & 33.2 & 65.83 & 0.698 \\
\bottomrule
\end{tabular}}
\caption{
\benchbg{pplcolor}{\strut Perplexity}, accuracy
(\benchbg{acccolor}{\strut HellaSwag}, \benchbg{acccolor}{\strut MMLU}),
bias
(\benchbg{othercolor}{\strut UnQover}, \benchbg{othercolor}{\strut BBQ}, \benchbg{othercolor}{\strut CP}),
and \benchbg{dtooverall}{\strut DTO} evaluation results for instruction-tuned and base models compressed using \textbf{Debias-SparseGPT} and baseline pruning methods under \texttt{1:4} sparsity.
CP = CrowS-Pairs. DTO = Distance-to-Optimum, computed using MMLU and UnQover.
$^{\dagger}$ indicates that the UnQover improvement of Debias-SparseGPT over SparseGPT is statistically significant under a paired $t$-test with $p < 0.01$.
}
\label{tab:debias-sparsegpt-newmodels}
\end{table*}

\begin{table*}[!t]
\centering
\resizebox{0.98\textwidth}{!}{%
\begin{tabular}{l l l P A A O O O D}
\toprule
\textbf{Regime}
& \textbf{Calibration Data}
& \textbf{Method}
& \textbf{PPL $\downarrow$}
& \textbf{HellaSwag $\uparrow$}
& \textbf{MMLU $\uparrow$}
& \textbf{UnQover $\uparrow$}
& \textbf{BBQ $\uparrow$}
& \textbf{CP $\downarrow$}
& \textbf{DTO $\downarrow$} \\
\midrule

\multicolumn{10}{c}{\textbf{Qwen-2.5-7B-Instruct}} \\
\midrule

Dense
& -
& -
& 7.14 & 56.99 & 68.71 & 73.17 & 86.30 & 60.82 & 0.291 \\

\cdashlinelr{1-10}

\multirow{4}{*}{\texttt{2:4}}
& \multirow{2}{*}{SS}
& SparseGPT
& 15.89 & 44.72 & 50.41 & 28.45 & 64.10 & 57.60 & 0.616 \\

&
& \textbf{Debias-SparseGPT}
& 16.17 & 44.53 & 48.16 & 24.94 & 66.70 & 57.72 & 0.645 \\

\cdashlinelr{2-10}

& \multirow{2}{*}{SS + UltraChat}
& SparseGPT
& 13.84 & 46.35 & 53.84 & 42.46 & 75.20 & 59.15 & 0.522 \\

&
& \textbf{Debias-SparseGPT}
& 13.37 & 46.54 & 54.17 & 47.26 & 78.60 & 56.65 & 0.494 \\

\bottomrule
\end{tabular}}
\caption{
\benchbg{pplcolor}{\strut Perplexity}, accuracy
(\benchbg{acccolor}{\strut HellaSwag}, \benchbg{acccolor}{\strut MMLU}),
bias
(\benchbg{othercolor}{\strut UnQover}, \benchbg{othercolor}{\strut BBQ}, \benchbg{othercolor}{\strut CP}),
and \benchbg{dtooverall}{\strut DTO} evaluation results for Qwen-2.5-7B-Instruct models compressed using SparseGPT and \textbf{Debias-SparseGPT} under structured \texttt{2:4} sparsity.
``SS'' denotes StereoSet, while ``SS + UltraChat'' denotes calibration using StereoSet augmented with 256 UltraChat examples.
CP = CrowS-Pairs. DTO = Distance-to-Optimum, computed using MMLU and UnQover; lower values indicate a better fairness-performance trade-off.
Overall, augmenting the calibration data with UltraChat improves the evaluation results under \texttt{2:4} sparsity, with the model compressed using Debias-SparseGPT achieving the lowest DTO among the pruned variants.
}
\label{tab:debias-sparsegpt-ultrachat-appendix}
\end{table*}

\begin{table*}[h!]
\centering
\resizebox{0.98\textwidth}{!}{%
\begin{tabular}{l P A A O O O}
\toprule
\textbf{Model + Method} 
& \textbf{PPL $\downarrow$} 
& \textbf{HellaSwag $\uparrow$} 
& \textbf{MMLU $\uparrow$}
& \textbf{UnQover $\uparrow$} 
& \textbf{BBQ $\uparrow$} 
& \textbf{CP $\downarrow$} \\
\midrule
\quad LLaMA-3.1-8B-Instruct (Dense)
& 6.99 & 57.49 & 63.17 & 30.10 & 76.40 & 61.72 \\
\cdashlinelr{1-7}
\quad \textit{Sparsity 0.25} \\
\quad + SparseGPT
& 7.31 & 57.04 & 62.74 & 27.30 & 71.30 & 61.42 \\
\quad + \textbf{Debias-SparseGPT}
& 7.32 & 56.96 & 63.10 & 34.90 & 72.30 & 60.58 \\

\cdashlinelr{1-7}
\quad \textit{Sparsity 0.50} \\
\quad + SparseGPT
& 11.26 & 51.09 & 47.69 & 27.41 & 49.70 & 60.52 \\
\quad + \textbf{Debias-SparseGPT}
& 11.34 & 50.92 & 48.32 & 28.60 & 46.20 & 63.74 \\

\cdashlinelr{1-7}
\quad \textit{Sparsity 1:4} \\
\quad + SparseGPT
& 8.17 & 55.56 & 59.11 & 35.60 & 67.10 & 60.64 \\
\quad + \textbf{Debias-SparseGPT}
& 8.19 & 55.45 & 59.76 & 60.46 & 70.70 & 59.81 \\

\cdashlinelr{1-7}
\quad \textit{Sparsity 2:4} \\
\quad + SparseGPT
& 29.36 & 39.13 & 27.06 & 31.60 & 43.10 & 60.29 \\
\quad + \textbf{Debias-SparseGPT}
& 30.86 & 39.39 & 24.57 & 31.60 & 44.80 & 61.30 \\

\midrule
\quad Vicuna-1.5-7B (Dense)
& 6.92 & 56.58 & 48.60 & 17.44 & 41.60 & 66.43 \\
\cdashlinelr{1-7}
\quad \textit{Sparsity 0.25} \\
\quad + SparseGPT
& 7.15 & 55.98 & 48.16 & 13.54 & 39.80 & 66.96 \\
\quad + \textbf{Debias-SparseGPT}
& 7.15 & 55.82 & 48.01 & 14.08 & 40.50 & 66.79 \\

\cdashlinelr{1-7}
\quad \textit{Sparsity 0.50} \\
\quad + SparseGPT
& 9.59 & 50.51 & 41.25 & 16.88 & 33.50 & 64.88 \\
\quad + \textbf{Debias-SparseGPT}
& 9.64 & 50.10 & 39.77 & 15.42 & 33.30 & 65.95 \\

\cdashlinelr{1-7}
\quad \textit{Sparsity 1:4} \\
\quad + SparseGPT
& 7.66 & 54.13 & 46.14 & 17.82 & 37.00 & 65.30 \\
\quad + \textbf{Debias-SparseGPT}
& 7.64 & 54.30 & 45.81 & 21.54 & 37.90 & 65.89 \\

\cdashlinelr{1-7}
\quad \textit{Sparsity 2:4} \\
\quad + SparseGPT
& 16.53 & 41.63 & 30.25 & 31.68 & 30.10 & 63.63 \\
\quad + \textbf{Debias-SparseGPT}
& 16.94 & 41.96 & 29.13 & 33.92 & 33.20 & 63.63 \\

\midrule
\quad Qwen-2.5-7B-Instruct (Dense)
& 7.14 & 56.99 & 68.71 & 73.17 & 86.30 & 60.82 \\

\cdashlinelr{1-7}
\quad \textit{Sparsity 0.25} \\
\quad + SparseGPT
& 7.36 & 56.32 & 68.46 & 72.35 & 86.80 & 59.81 \\
\quad + \textbf{Debias-SparseGPT}
& 7.37 & 56.44 & 68.39 & 73.43 & 86.60 & 60.05 \\

\cdashlinelr{1-7}
\quad \textit{Sparsity 0.50} \\
\quad + SparseGPT
& 9.59 & 52.66 & 62.18 & 78.35 & 82.50 & 58.80 \\
\quad + \textbf{Debias-SparseGPT}
& 9.67 & 52.00 & 62.59 & 80.35 & 82.50 & 60.64 \\

\cdashlinelr{1-7}
\quad \textit{Sparsity 1:4} \\
\quad + SparseGPT
& 7.92 & 55.15 & 67.35 & 70.60 & 85.40 & 61.24 \\
\quad + \textbf{Debias-SparseGPT}
& 7.92 & 55.06 & 67.73 & 74.41 & 86.60 & 59.99 \\

\cdashlinelr{1-7}
\quad \textit{Sparsity 2:4} \\
\quad + SparseGPT
& 15.89 & 44.72 & 50.41 & 28.45 & 64.10 & 57.60 \\
\quad + \textbf{Debias-SparseGPT}
& 16.17 & 44.53 & 48.16 & 24.94 & 66.70 & 57.72 \\

\bottomrule
\end{tabular}}
\caption{
Perplexity, downstream accuracy, and bias evaluation results for all models across different sparsity regimes, including dense models and models pruned with SparseGPT and Debias-SparseGPT under unstructured and semi-structured pruning settings. 
Overall, models compressed with Debias-SparseGPT generally achieve higher UnQover accuracy than models compressed with SparseGPT while maintaining comparable perplexity and downstream accuracy.
The largest improvement is observed for LLaMA under \texttt{1:4} sparsity, where UnQover accuracy increases from 35.60\% to 60.46\%.
}
\label{tab:debias-sparsegpt-all-models-appendix}
\end{table*}

\paragraph{Calibration Under 2:4 Sparsity}
\autoref{tab:debias-sparsegpt-ultrachat-appendix} reports evaluation results for models compressed using a mixture of UltraChat data \citep{ding-etal-2023-enhancing} and StereoSet data \citep{nadeem-etal-2021-stereoset}, compared to a StereoSet-only baseline under the 2:4 sparsity setting for the Qwen model.
Under 2:4 sparsity with StereoSet-only calibration, we observe a substantial decrease in model performance, particularly on MMLU and UnQover. 
We find that augmenting the calibration data with 256 UltraChat examples improves the performance of models compressed with both methods. 
The model compressed with Debias-SparseGPT, using the augmented calibration set, achieves 54.17\% on MMLU and 47.26\% on UnQover, compared with 48.16\% and 24.94\%, respectively, when StereoSet alone is used. The corresponding DTO decreases from 0.645 to 0.494. Overall, the score improvement is larger for the model compressed with Debias-SparseGPT than for the model compressed with SparseGPT, with MMLU accuracy increasing from 48.16\% to 54.17\% compared with 50.41\% to 53.84\%, and UnQover accuracy increasing from 24.94\% to 47.26\% compared with 28.45\% to 42.46\%.

\begin{table*}[h]
\centering
\small
\resizebox{0.8\textwidth}{!}{
\begin{tabular}{l P A O D}
\toprule
\textbf{Calibration Data} & \textbf{PPL $\downarrow$} & \textbf{MMLU $\uparrow$} & \textbf{UnQover $\uparrow$} & \textbf{DTO $\downarrow$} \\
\midrule
SS + UltraChat (4)    & 16.56 & 51.28 & 32.60 & 0.588 \\
SS + UltraChat (16)   & 16.22 & 47.43 & 40.36 & 0.562 \\
SS + UltraChat (64)   & 15.28 & 52.50 & \textbf{50.32} & \textbf{0.486} \\
\textbf{SS + UltraChat (256)} & \textbf{13.37} & \textbf{54.17} & 47.26 & 0.494 \\
SS + UltraChat (1024) & 13.65 & 51.89 & 48.04 & 0.501 \\
\bottomrule
\end{tabular}}
\caption{Effect of UltraChat calibration size on the performance of Qwen-2.5-7B-Instruct compressed with Debias-SparseGPT under 2:4 semi-structured sparsity. Each row corresponds to a different number of UltraChat calibration examples (4, 16, 64, 256, 1024). We report WikiText-2 perplexity, zero-shot MMLU accuracy, UnQover accuracy, and the resulting DTO score.}
\label{tab:ultrachat_scaling}
\end{table*}

\begin{table*}[h]
\centering
\resizebox{0.98\textwidth}{!}{%
\begin{tabular}{l P A O O O O O D}
\toprule
\textbf{Calibration Data} 
& \textbf{PPL $\downarrow$} 
& \textbf{MMLU $\uparrow$} 
& \multicolumn{5}{>{\columncolor{othercolor}}c}{\textbf{UnQover $\uparrow$}}
& \textbf{DTO $\downarrow$} \\
& 
& 
& \textbf{Religion}
& \textbf{Nationality}
& \textbf{Race}
& \textbf{Gender}
& \textbf{Avg.}
& \\
\midrule
StereoSet (Gender)     
& 13.56 & 56.89 & 34.50 & 22.30 & 32.49 & 55.49 & 36.20 & 0.544 \\
StereoSet (Race)     
& 13.95 & \textbf{57.42} & 45.30 & 25.70 & 38.61 & 55.15 & 41.20 & 0.513 \\
StereoSet (Religion) 
& 13.64 & 56.96 & \textbf{57.68} & \textbf{40.74} & \textbf{53.03} & 58.53 & \textbf{52.50} & \textbf{0.453} \\
\midrule
CrowS-Pairs (All)      
& 14.68 & 53.09 & 41.17 & 27.88 & 37.92 & 52.21 & 39.80 & 0.540 \\
StereoSet (All)      
& \textbf{13.37} & 54.17 & 44.05 & 39.96 & 45.61 & \textbf{59.42} & 47.26 & 0.494 \\
\bottomrule
\end{tabular}}
\caption{
Effect of category-specific calibration data on the performance of Qwen-2.5-7B-Instruct compressed with Debias-SparseGPT under \texttt{2:4} semi-structured sparsity.
We report results for models compressed using calibration data composed of gender-, race-, and religion-specific StereoSet pairs, full StereoSet, and full CrowS-Pairs benchmarks, combined with the same 256 UltraChat examples used in \autoref{tab:ultrachat_scaling}.
We report WikiText-2 perplexity, zero-shot MMLU accuracy, UnQover category-level accuracies, the overall UnQover average, and the resulting DTO score.
}
\label{tab:stereoset_category_calibration}
\end{table*}

\section{Calibration Data Ablation Experiments}
\label{app:calibration-data-choice-rebuttal}

In this appendix, we provide additional details on the calibration data used for model compression and present calibration data ablation experiments.

\paragraph{Choice of Calibration Data}

The Debias-SparseGPT objective defined in Eq.~\eqref{eq:fair-sparse-objective} requires paired pro- and anti-stereotypical inputs to construct the bias-aware Hessian in Eq.~\eqref{eq:debias-hessian}. 
We therefore use minimal contrastive pairs from StereoSet \citep{nadeem-etal-2021-stereoset}, which provide paired stereotypical and anti-stereotypical sentences. 
This choice is consistent with prior debiasing work that uses StereoSet for evaluating or constructing debiasing interventions, including CDA, INLP, SentenceDebias, Self-Debias, and BiasEdit \citep{xu-etal-2025-biasedit,meade-etal-2022-empirical}. 
To study the effect of calibration data, we perform two ablations: first, we vary the number of added UltraChat examples, and second, we compare category-specific StereoSet calibration with full StereoSet and CrowS-Pairs calibration.
In this appendix, we experiment with the Qwen-2.5-7B model under the \texttt{2:4} sparsity regime.

\paragraph{UltraChat Ablation}

We report evaluation results for models compressed with UltraChat in \autoref{tab:ultrachat_scaling}. 
We find that, under \texttt{2:4} semi-structured sparsity, adding a moderate number of UltraChat examples substantially improves the fairness-performance trade-off. 
With four additional UltraChat examples, the model compressed with Debias-SparseGPT achieves better general and fairness performance (51.28\% MMLU and 32.60\% UnQover accuracy) compared to the StereoSet-only calibration setting. 
Increasing the number of UltraChat examples to 256 yields the lowest perplexity (13.37) and highest MMLU accuracy (54.17\%) among the evaluated calibration sizes, with an UnQover accuracy of 47.26\% and DTO of 0.494. The 64-example setting yields the highest average UnQover accuracy (50.32\%) and lowest DTO (0.486).
Adding more UltraChat examples does not further improve the trade-off. 
With 1024 UltraChat examples, perplexity increases slightly to 13.65, MMLU decreases to 51.89\%, UnQover decreases to 48.04\%, and DTO increases to 0.501. 
Overall, these results suggest that contextually rich calibration data could further improve the performance of compressed models under aggressive \texttt{2:4} sparsity.

\paragraph{StereoSet and CrowS-Pairs Calibration Ablation}

Next, we analyze the performance of models compressed using category-specific subsets of the StereoSet calibration data. For each group category $\mathcal{G} \in \{\text{gender}, \text{race}, \text{religion}\}$, we construct a calibration subset consisting of the corresponding minimal contrastive pairs from StereoSet. We report the category-specific calibration ablation results in \autoref{tab:stereoset_category_calibration}.
For these experiments, we use Debias-SparseGPT under \texttt{2:4} semi-structured sparsity and combine each calibration setting with the same 256 UltraChat examples used in \autoref{tab:ultrachat_scaling} to ensure a consistent comparison. We compare gender-, race- and religion-specific StereoSet subsets with the full StereoSet and full CrowS-Pairs calibration sets.
We find that, among the category-specific StereoSet subsets, religion-specific calibration yields the highest average UnQover accuracy and the lowest DTO.
Models compressed using the religion-specific subset achieve 56.96\% MMLU, 52.50\% average UnQover, and a DTO of 0.453, compared with 56.89\% MMLU, 36.20\% average UnQover, and a DTO of 0.544 when using the gender-specific subset. Religion-specific calibration also yields the highest category-level UnQover scores among the category-specific subsets.

We further find that calibration using CrowS-Pairs results in lower performance than using the full StereoSet calibration data, with higher perplexity (14.68 vs.\ 13.37), lower MMLU accuracy (53.09\% vs.\ 54.17\%), lower average UnQover accuracy (39.80\% vs. 47.26\%), and higher DTO (0.540 vs. 0.494). 
Overall, full StereoSet calibration outperforms CrowS-Pairs across all reported metrics, while religion-specific StereoSet calibration achieves the strongest fairness-performance trade-off among the evaluated paired calibration sets.

Taken together, the calibration data ablation experiments show that the choice of paired calibration data affects the fairness-performance trade-off. Among the category-specific StereoSet subsets, religion-calibration achieves the highest average UnQover accuracy and the lowest DTO, while full StereoSet calibration yields the lowest perplexity and outperforms CrowS-Pairs across all reported metrics.
We also observe that the number of UltraChat examples affects compressed-model performance. Among the evaluated calibration sizes, 256 UltraChat examples yield the lowest perplexity and the highest MMLU accuracy, whereas 64 examples yield the highest UnQover accuracy and the lowest DTO. The category-specific experiments also provide evidence of cross-category transfer on UnQover. Religion-specific calibration yields higher average UnQover accuracy than full StereoSet calibration, whereas gender- and race-specific calibration yield lower average accuracy.

\begin{table*}[!t]
    \centering
    \begin{tabular}{l O O}
        \toprule
        \textbf{Model + Method}
        & \textbf{RealToxicityPrompts} $\downarrow$
        & \textbf{HarmBench} $\downarrow$ \\
        & \textbf{Unsafe Rate}
        & \textbf{Unsafe Rate} \\
        \midrule
        LLaMA-3.1-8B + SparseGPT
        & 0.033 & 0.265 \\
        LLaMA-3.1-8B + \textbf{Debias-SparseGPT}
        & \textbf{0.028} & \textbf{0.190} \\
        \midrule
        Vicuna-1.5-7B + SparseGPT
        & 0.344 & 0.510 \\
        Vicuna-1.5-7B + \textbf{Debias-SparseGPT}
        & \textbf{0.331} & \textbf{0.425} \\
        \midrule
        Qwen-2.5-7B + SparseGPT
        & 0.035 & 0.055 \\
        Qwen-2.5-7B + \textbf{Debias-SparseGPT}
        & \textbf{0.031} & \textbf{0.025} \\
        \bottomrule
    \end{tabular}
    \caption{
        Toxicity and safety evaluation of models compressed with SparseGPT and Debias-SparseGPT under 1:4 semi-structured sparsity.
        We report the proportion of generated responses classified as unsafe by LLaMA Guard 3-1B; lower values are better.
        Debias-SparseGPT consistently yields lower unsafe-response rates across all evaluated model families and benchmarks.
    }
    \label{tab:safety-evaluation}
\end{table*}

\section{Safety Evaluation}\label{app:safety-evaluation}

To complement our evaluation of representational bias, we additionally assess whether Debias-SparseGPT affects broader model safety.
First, we evaluate generations from compressed models conditioned on inputs from the \textbf{RealToxicityPrompts} benchmark~\citep{gehman2020realtoxicityprompts}, following prior work studying the impact of model compression on toxicity and safety~\citep{xu2022can,xu-etal-2024-beyond-perplexity,hong2024decoding}.
RealToxicityPrompts contains human-written prompts designed to elicit potentially toxic model continuations. 
We randomly sample 1,200 prompts for evaluation and use the benchmark implementation provided by the LM Evaluation Harness framework.\footnote{\href{https://github.com/EleutherAI/lm-evaluation-harness}{github.com/EleutherAI/lm-evaluation-harness}}
To evaluate the generated continuations, we classify outputs using the LLaMA Guard 3-1B model~\citep{inan2023llama}.\footnote{The \href{https://www.perspectiveapi.com/}{Perspective API}, used in the source implementation, no longer accepts new usage requests.}
  
Next, we evaluate model safety on \textbf{HarmBench}~\citep{mazeika2024harmbench}, a standardized benchmark containing prompts related to cybercrime, the production or use of chemical and biological weapons or drugs, misinformation, harassment, illegal activities, and other risks to human well-being.

For both benchmarks, we report the proportion of generated responses classified as unsafe by the Guard model; lower values are better.
Generation is performed using deterministic greedy decoding with temperature set to zero.
We compare the performance of the models compressed using both methods under the 1:4 semi-structured sparsity setting. 

We report the results in \autoref{tab:safety-evaluation}.
We find that Debias-SparseGPT achieves lower unsafe-response rates than SparseGPT-compressed models across all three model families on both RealToxicityPrompts and HarmBench. 
The largest absolute reduction is observed for Vicuna-1.5-7B on HarmBench, where the unsafe-response rate decreases from $0.510$ to $0.425$. For LLaMA-3.1-8B, the rate decreases from $0.265$ to $0.190$, while Qwen-2.5-7B exhibits the lowest HarmBench unsafe-response rate overall, decreasing from $0.055$ to $0.025$. 

These results suggest that incorporating the proposed bias-aware objective during pruning does not introduce an adverse safety trade-off and, in the evaluated settings, is associated with improved performance relative to SparseGPT baselines.

\section{Analysis of Learned Sparsity Patterns}
\label{app:sparsity-pattern-analysis}

To examine how the bias-aware Hessian affects pruning decisions, we compare Qwen-2.5-7B-IT models compressed with SparseGPT and Debias-SparseGPT under 2:4 semi-structured sparsity. 
Both models are compressed using the same calibration data, consisting of full StereoSet and 256 examples from UltraChat. 
We analyze all seven pruned matrices in each layer: \texttt{q\_proj}, \texttt{k\_proj}, \texttt{v\_proj}, and \texttt{o\_proj} in the attention module, and \texttt{gate\_proj}, \texttt{up\_proj}, and \texttt{down\_proj} in the MLP module.

For each final pruned matrix, we extract a binary mask in which a value of 1 denotes a pruned weight. We then compute four measures: 1) weight disagreement, the fraction of matrix positions where pruning decisions differ between the two masks, where larger values indicate greater mask divergence; 2) \texttt{2:4} pattern disagreement, the fraction of consecutive four-weight groups for which the two masks select different zero positions; 3) pruned-index Jaccard, the intersection-over-union of the two sets of pruned indices, where values closer to 1 indicate more similar masks and values closer to 0 indicate less overlap; and 4) relative Frobenius difference, computed as
$\|\hat{\mW}^{(1)}-\hat{\mW}^{(0)}\|_F/
\|\hat{\mW}^{(0)}\|_F$, which measures the relative difference between the final pruned weight matrices. 
Together, these measures reflect the differences in the final pruning masks between models compressed with SparseGPT and Debias-SparseGPT.

\autoref{tab:sparsity-pattern-analysis} reports the results for matrices with the largest 2:4 pattern disagreement.
We find that among the 20 matrices with the largest 2:4 pattern disagreement, 16 correspond to attention output projections, indicating that the effect of the bias-aware Hessian is non-uniform across layers and modules and is concentrated primarily in \texttt{self\_attn.o\_proj}.

\begin{table*}[t]
\centering
\setlength{\tabcolsep}{5pt}
\begin{tabular}{clcccc}
\toprule
\textbf{Layer} &
\textbf{Module} &
\textbf{Weight} &
\textbf{2:4 Pattern} &
\textbf{Pruned-index} &
\textbf{Relative Frobenius} \\
&
&
\textbf{Disagreement} &
\textbf{Disagreement} &
\textbf{Jaccard} &
\textbf{Difference} \\
\midrule
26 & \texttt{self\_attn.o\_proj} & \textbf{0.0975} & \textbf{0.1928} & \textbf{0.8223} & 0.3040 \\
27 & \texttt{self\_attn.o\_proj} & 0.0909 & 0.1797 & 0.8333 & \textbf{0.3121} \\
22 & \texttt{self\_attn.o\_proj} & 0.0831 & 0.1644 & 0.8465 & 0.2731 \\
19 & \texttt{self\_attn.o\_proj} & 0.0829 & 0.1643 & 0.8468 & 0.2636 \\
13 & \texttt{self\_attn.o\_proj} & 0.0824 & 0.1633 & 0.8477 & 0.2602 \\
9  & \texttt{self\_attn.o\_proj} & 0.0817 & 0.1618 & 0.8489 & 0.2684 \\
18 & \texttt{self\_attn.o\_proj} & 0.0812 & 0.1606 & 0.8499 & 0.2670 \\
8  & \texttt{self\_attn.o\_proj} & 0.0802 & 0.1586 & 0.8515 & 0.2617 \\
12 & \texttt{self\_attn.o\_proj} & 0.0791 & 0.1569 & 0.8534 & 0.2609 \\
7  & \texttt{self\_attn.o\_proj} & 0.0792 & 0.1564 & 0.8533 & 0.2680 \\
0  & \texttt{mlp.up\_proj}      & 0.0781 & 0.1543 & 0.8552 & 0.2911 \\
21 & \texttt{self\_attn.o\_proj} & 0.0774 & 0.1534 & 0.8563 & 0.2490 \\
16 & \texttt{self\_attn.o\_proj} & 0.0771 & 0.1529 & 0.8568 & 0.2543 \\
23 & \texttt{self\_attn.o\_proj} & 0.0749 & 0.1486 & 0.8607 & 0.2485 \\
0  & \texttt{mlp.gate\_proj}    & 0.0750 & 0.1484 & 0.8604 & 0.2785 \\
14 & \texttt{mlp.down\_proj}    & 0.0746 & 0.1482 & 0.8612 & 0.2525 \\
25 & \texttt{self\_attn.o\_proj} & 0.0742 & 0.1473 & 0.8618 & 0.2481 \\
20 & \texttt{self\_attn.o\_proj} & 0.0738 & 0.1463 & 0.8625 & 0.2513 \\
13 & \texttt{mlp.down\_proj}    & 0.0735 & 0.1461 & 0.8631 & 0.2515 \\
1  & \texttt{self\_attn.o\_proj} & 0.0738 & 0.1457 & 0.8626 & 0.2662 \\
\bottomrule
\end{tabular}
\caption{
Matrices with the largest 2:4 pruning-pattern disagreement between SparseGPT and Debias-SparseGPT for the Qwen-2.5-7B-IT model.
Both models are compressed under 2:4 semi-structured sparsity using the same StereoSet and UltraChat calibration data.
Layer indices are zero-based and range from 0 (first layer) to 32 (last layer).
Larger disagreement values indicate stronger changes in pruning decisions induced by the bias-aware Hessian, whereas larger pruned-index Jaccard values indicate greater mask overlap.
Bold values indicate the largest disagreement or relative-difference values and the smallest pruned-index Jaccard value.
}
\label{tab:sparsity-pattern-analysis}
\end{table*}

\end{document}